\documentclass[11pt]{article}

\usepackage[final]{acl}

\usepackage{times}
\usepackage{latexsym}

\usepackage[T1]{fontenc}
\usepackage[utf8]{inputenc}

\usepackage{microtype}

\usepackage{inconsolata}

\usepackage{graphicx}

\usepackage[position=bottom]{subfig}
\usepackage{makecell}
\usepackage{titletoc} %目录包
\usepackage{xcolor}
\definecolor{bg_gray}{RGB}{245,245,245}
\usepackage{enumitem}
\usepackage{geometry}
\usepackage{amsmath}
\usepackage[most]{tcolorbox} % 用于漂亮的盒子
\definecolor{keywordcolor}{RGB}{0, 50, 150} % 深蓝色
\definecolor{bracketcolor}{RGB}{100, 100, 100} % 灰色
\usepackage{hyperref}       % hyperlinks
\usepackage{url}            % simple URL typesetting
\usepackage{booktabs}       % professional-quality tables
\usepackage{amsfonts}       % blackboard math symbols
\usepackage{nicefrac}       % compact symbols for 1/2, etc.
\usepackage{microtype}      % microtypography
\usepackage[table,xcdraw]{xcolor}
\usepackage{enumitem}
\usepackage{colortbl}
\usepackage{threeparttable}
\usepackage{multirow}
\usepackage{pifont} 
\usepackage{graphicx}
\usepackage{amsmath} 
\usepackage{stfloats} 
\usepackage{caption}
\usepackage{subcaption}
\usepackage{booktabs} % 更好的表格线
\usepackage[normalem]{ulem}
\def\eg{\emph{e.g.}}

\newcommand{\ptag}[1]{\textcolor{keywordcolor}{$\langle$\textit{#1}$\rangle$}}

\newcommand{\tablestyle}[2]{\setlength{\tabcolsep}{#1}\renewcommand{\arraystretch}{#2}\centering\footnotesize}
\definecolor{highlightyellow}{rgb}{1.0, 0.98, 0.7} % 淡黄色用于高亮行
\definecolor{headergray}{rgb}{0.92, 0.92, 0.92}   % 淡灰色用于分节标题
\newcommand{\RNum}[1]{\uppercase\expandafter{\romannumeral #1\relax}}
\title{\raisebox{-0.15cm}{\includegraphics[height=1.3em]{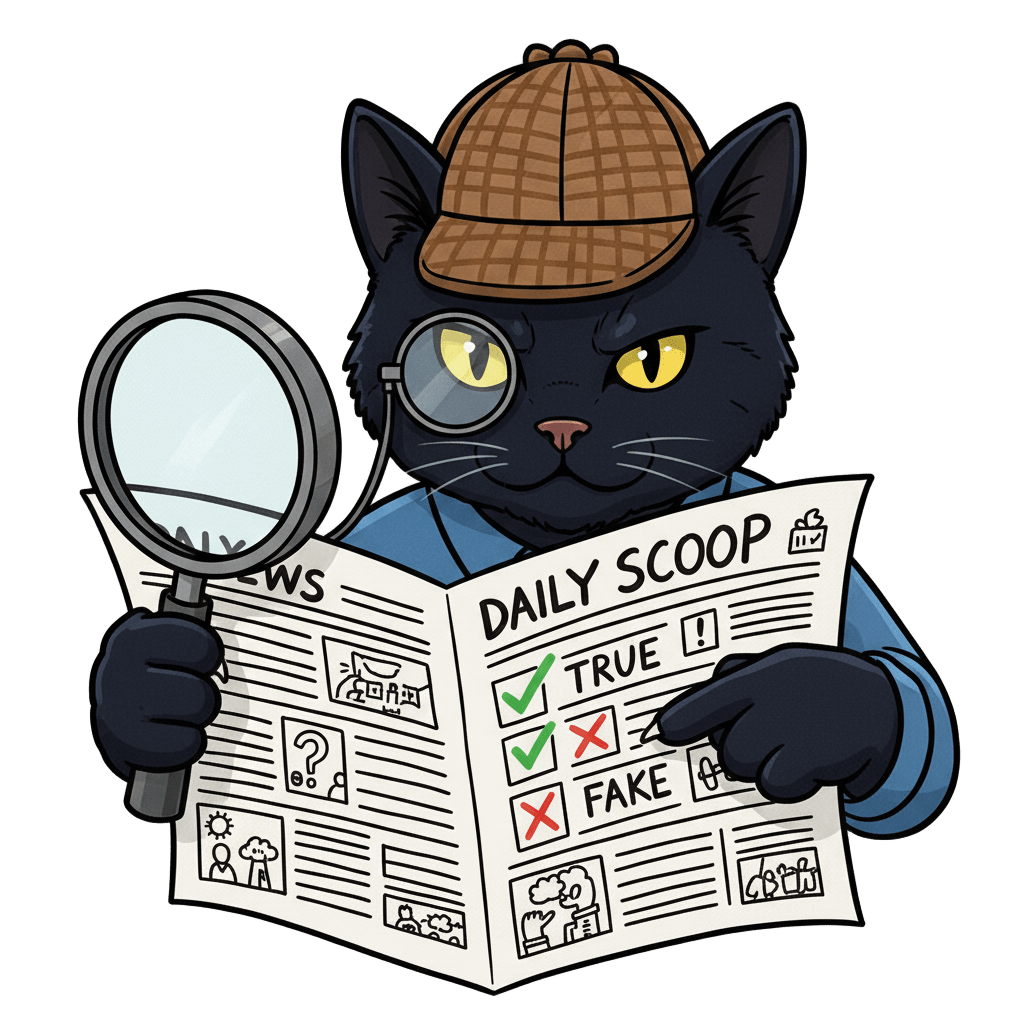}}~Cultivating Forensic Reasoning for Generalizable Multimodal Manipulation Detection}

\author{
\textbf{Yuchen Zhang}$^{\textbf{1}}$ \quad
\textbf{Yaxiong Wang}$^{\textbf{2}}$\thanks{Corresponding author.} \quad
\textbf{Kecheng Han}$^{\textbf{1}}$ \quad
\textbf{Yujiao Wu}$^{\textbf{3}}$ \\
\textbf{Lianwei Wu}$^{\textbf{4}}$ \quad
\textbf{Li Zhu}$^{\textbf{1}}$ \quad
\textbf{Zhedong Zheng}$^{\textbf{5}}$\\
$^{1}$School of Software Engineering, Xi'an Jiaotong University \\
$^{2}$School of Computer Science and Information Engineering, Hefei University of Technology \\ 
$^{3}$CSIRO \quad
$^{4}$Northwestern Polytechnical University \quad
$^{5}$University of Macau\\
{\tt\small yczhang@stu.xjtu.edu.cn} \quad
{\tt\small wangyx@hfut.edu.cn}\\
}

\begin{document}
\maketitle

\begin{abstract}
Recent advances in generative AI have significantly enhanced the realism of multimodal media manipulation, thereby posing substantial challenges to manipulation detection. Existing manipulation detection and grounding approaches predominantly focus on manipulation type classification under result-oriented supervision, which not only lacks interpretability but also tends to overfit superficial artifacts. In this paper, we argue that generalizable detection requires incorporating explicit forensic reasoning, rather than merely classifying a limited set of manipulation types, which fails to generalize to unseen manipulation patterns.
To this end, we propose \textbf{REFORM}, a reasoning-driven framework that shifts learning from outcome fitting to process modeling. REFORM adopts a three-stage curriculum that first induces forensic rationales, then aligns reasoning with final judgments, and finally refines logical consistency via reinforcement learning. To support this paradigm, we introduce \textbf{ROM}, a large-scale dataset with rich reasoning annotations. Extensive experiments show that REFORM establishes new state-of-the-art performance with superior generalization, achieving 81.52\% ACC on ROM, 76.65\% ACC on DGM4, and 74.9 F1 on MMFakeBench. The code is available at \url{https://github.com/YcZhangSing/REFORM}.
\end{abstract}
\section{Introduction}

% 第一段，背景引入
%%%AI/diffusion发展，多媒体内容丰富，但虚假篡改 恶意操纵内容影响广泛切恶劣，亟需高效实用的检测模型.....
% \yxnote{Background and task definition....}
The democratization of Generative AI \citep{sd_v3_5_web,QwenVL,LLAVA_NeurIPS}, has precipitated a paradigm shift in digital content creation. While this technological renaissance empowers creativity, it simultaneously lowers the barrier for fabricating hyper-realistic misinformation. The proliferation of sophisticated multimodal manipulations, ranging from subtle face swaps to fully synthesized news events, poses severe threats to information integrity and public trust \citep{pan_EMNLP2023_risk,park_EMNLP2022_challenges,yaxiongAIGCDet3}. Consequently, detecting and grounding multimodal media manipulation (DGM4) has attracted wide attention and seen notable progress these years~\cite{DGM4_TPAMI,zhang2025asap,liu2024fkaowl}.
%developing robust detection systems capable of identifying and interpreting these forgeries has become an urgent priority in the multimedia security community.

\begin{figure}[t]
    \centering
    \vspace{-.1in}
    \includegraphics[width=\columnwidth]{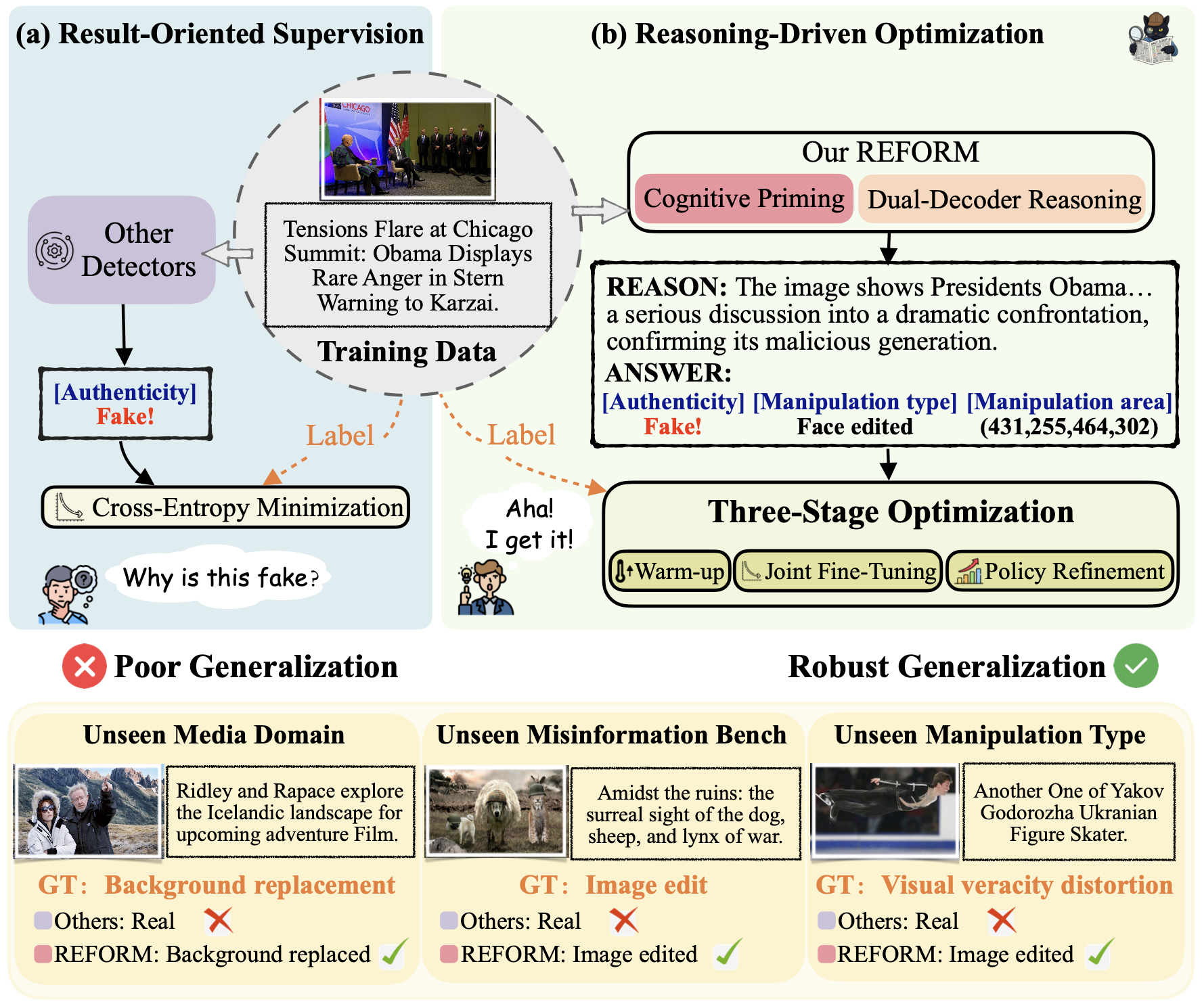}  
    \vspace{-0.6cm}
    \caption{Comparison between learning paradigms. (a) The prevailing \textit{Result-Oriented Supervision} usually suffers from poor generalization by merely fitting statistical artifacts of training data. (b) Our \textit{Reasoning-Driven Optimization} facilitates robust generalization by explicitly optimizing the forensic reasoning chain, enabling the model to uncover intrinsic inconsistencies effectively across unseen domains.}
    \vspace{-0.5cm}
    \label{fig:motivation}
\end{figure}

%第二段，引出关注的点：泛化性 
% \yxnote{The generalization ability of detection models play a vital role in practical application scenarios across diverse data domains. Many researchers have developed methods to enhance the generalizaliblity. FKA-OWL..... AMD ..... }
%pioneering work of DGM4 \citep{DGM4_TPAMI} established the foundational benchmark for multimodal manipulation detection, catalyzing a wave of subsequent studies \citep{SAMM_shenjinjie_MM25,trusvl_emnlp_2025}. 
Despite the progress made on the DGM4 problem, a pivotal challenge still remains: \textit{generalization}. In practical scenarios, detectors must operate on diverse news domains and unseen manipulation patterns. Prior research has largely focused on invariant feature learning and knowledge augmentation~\cite{liu2024fkaowl,AMD_MDSM,zhang2025asap}. For example, FKA-Owl \citep{liu2024fkaowl} augments Large Vision-Language Models with semantic correlations of forgery knowledge and visual artifacts for generalization enhancement. Concurrently, AMD \citep{AMD_MDSM} explicitly aligns visual artifacts with textual inconsistencies in a shared latent space to learn invariant forensic representations robust to semantic contexts. While effective, these methods generally optimize models to map multimodal news directly to label annotations. This paradigm, which is heavily reliant on \textbf{result-oriented supervision}, tends to encourage models to fit specific statistical artifacts rather than cultivating a deep, transferable understanding of the underlying forensic logic.

%第三段，指出问题，引出我们的方法，
%我们motivation切 更加侧重 推理能力的过程学习，而不是基于监督学习的结果拟合
%这种推理能力可以提高泛化性，比简单进行结果拟合更好泛化
%通过这个思路，把reasoning和RL串起来
% \yxnote{Despite these progress, the above results-oriented supervised methods tends to predict the supervision label, which holds higher risk to overfitting. In contrast, we argue that the reasoning ability of tracing the manipulation among the multimedia is the key of generalization. Motivated by this, this paper targets to cultivate the reasoning ability of the detection and grounding models instead of using result prediction-based learning paradigm. To this end, we propose a three stage learning framework. }
As illustrated in Fig. \ref{fig:motivation} (a), relying solely on outcome supervision struggles to equip models with the intrinsic logic required for robust generalization. The essential reason is that  optimizing exclusively for the final prediction (i.e., "Is this fake?") does not guarantee that the model implicitly learns the underlying forensic rationale, often leading to performance degradation when facing novel forgeries~\citep{SFT_RL_ICML25,reason_cot_ACL}. 
To address this, we argue that the core of generalizable detection lies in explicitly cultivating forensic reasoning capabilities, as depicted in Fig.\ref{fig:motivation} (b). Akin to a human forensic analyst, a model should not merely memorize the appearance of a forgery but rather deduce why an image-text pair is inconsistent through a logical chain of evidence. However, simply supervising this reasoning process is insufficient. While Supervised Fine-Tuning establishes a basic alignment, it remains constrained by the passive imitation of static annotations \citep{SFT_RL_ICML25,S2R_RL_SFT_ACL25}. 
%In contrast, Reinforcement Learning is more ininclided to causal relationship learning, thus facilitating \textbf{Reasoning-Driven Optimization}. This superior paradigm enables the model to actively explore and refine its logical paths to ensure the derived reasoning strictly supports the final verdict.

%第四段，简单概览方法
Motivated by this insight, we propose a new framework that shifts the learning objective from outcome fitting to \textbf{Reasoning-Driven Optimization}. To this end, we propose \textbf{R}easoning-\textbf{E}nhanced \textbf{F}orensic \textbf{O}ptimization via \textbf{R}einforcement \textbf{M}odeling (\textbf{REFORM}),
a reasoning-endowed architecture designed to detect, ground, and explain multimodal manipulations. 
% \zycnote{Structurally, REFORM employs a novel encoder-dual-decoder design that decouples the generation of linguistic forensic reasoning from the final result prediction. This ensures logical consistency while allowing the reasoning branch to be bypassed for rapid, accuracy-preserving inference.}
To effectively cultivate this capability, we implement a progressive three-stage learning curriculum. First, we perform \textit{Cognitive Reasoning Warm-up}, teaching the model to articulate forensic rationales via data distillation. Second, we execute \textit{Reasoning-Endowed Joint Fine-Tuning}, where the model learns to align its final judgment with its reasoning chain. Finally, recognizing that supervised learning suffers from exposure bias and lacks the ability to self-correct, we introduce \textit{Constraint-Aware Policy Refinement}. Leveraging Reinforcement Learning with Group Relative Policy Optimization \citep{DeepSeekMath_GRPO}, we incentivize the model to explore optimal reasoning paths that are logically consistent with the final verdict. This optimization strictly constrains the generation process with forensic accuracy, enabling REFORM to internalize robust judgment logic rather than fitting domain-specific patterns.

%第五段，总结
To rigorously benchmark this new paradigm, we construct  
% \textbf{R}easoning-\textbf{E}nhanced \textbf{A}nalysis of \textbf{L}LM-driven \textbf{M}anipulation (\textbf{REALM}) 
\textbf{R}easoning-enhanced analysis for \textbf{O}mnibus \textbf{M}anipulation (\textbf{ROM}) dataset. 
% \yxnote{LLM here cannot summary the full features of our new dataset, what about ``Reasoning-enhanced analysis for Omnibus Manipulation (ROM)'' dataset? }
%Extending 
Beyond the face-centric scope of previous benchmarks \citep{AMD_MDSM,DGM4,jingchun_deepfake}, ROM introduces scene-level synthesis and, crucially, provides detailed reasoning annotations for over \textbf{704k} samples to support process-oriented learning.
In summary, our contributions are threefold:
\begin{itemize}[leftmargin=*]
    \item We identify the limitations of result-oriented manipulation detection and propose a paradigm shift towards reasoning-driven analysis, implementing REFORM, a new framework for robust multimodal manipulation detection, grounding, and forensic explanation.
    % a new encoder-dual-decoder architecture \yxnote{Dual encoder was not mentioned before, so, suddenly occurring here is not smooth,} that integrates detection, grounding, and forensic explanation.
    \item We introduce a progressive three-stage training framework culminating in a GRPO-based RL phase. This approach realizes Reasoning-Driven Optimization, effectively aligning the model's cognitive process with forensic logic to distinguish intrinsic anomalies from statistical biases.
    \item We curate ROM, a large-scale, comprehensive benchmark that includes omnibus manipulation beyond face-related manipulations and provides high-quality reasoning supervision, setting a new standard for interpretable multimodal forensics.
\end{itemize}

\section{Related Work}
\label{sec:related_work}

\noindent\textbf{Multimodal Misinformation Detection.} %The DGM4 task~\citep{DGM4_TPAMI} addresses sophisticated aligned forgeries. 
Given challenging aligned forgeries~\citep{DGM4_TPAMI}, recent approaches typically leverage external knowledge or refine internal features. For instance, FKA-Owl~\citep{liu2024fkaowl} utilizes LVLMs for factual verification, while RamDG~\citep{SAMM_shenjinjie_MM25} cross-references news with attribute databases. Conversely, feature-centric methods, \eg, AMD~\citep{AMD_MDSM}, encode artifact tokens alongside semantic content. However, these methods predominantly rely on \textit{result-oriented supervision}. This paradigm usually encourages shortcut learning~\citep{reason_cot_ACL}, causing models to overfit statistical artifacts rather than logical evidence, which degrades generalization on unseen domains~\citep{SFT_RL_ICML25}.

\noindent\textbf{Vision Language Models and Reasoning Optimization.} Large Vision Language Models, such as Qwen-VL~\citep{bai2025qwen25} and InternVL~\citep{wang2025internvl3}, have revolutionized multimodal understanding and been adapted for forensics via Supervised Fine-Tuning (SFT)~\citep{trust_VL_EMNLP,FakeSV_EMNLP25_yxw,yaxiongAIGCDet4}. However, SFT often fails to cultivate robust reasoning as it focuses on mimicking outputs, limiting generalization and self-correction~\citep{LaV_CoT_huang2025,SFT_RL_ICML25}. To address this, \textit{Reasoning-Driven Optimization} utilizing Chain-of-Thought~\citep{Self_Rewarding_25,reason_cot_ACL} and Reinforcement Learning methods like GRPO~\citep{DeepSeekMath_GRPO} has emerged to incentivize the reasoning process itself~\citep{So_Fake_R1,Parallel_R1}. 
Inspired by this, our REFORM framework integrates RL-based optimization to enforce logical consistency between forensic reasoning chains and verdicts, enabling robust and generalizable detection.
% 可以将“推理与最终决策不一致”作为训练信号（判别推理是否充分支撑结论），从而约束模型在内部依赖真实证据
\section{Methodology}
%我们的pipeline分为两步，第一步是reason数据蒸馏
% Our pipeline consists of two phases: Data Preparation and AMDv2.

\subsection{Data Preparation}

\begin{figure*}[t]
%%原图编译太慢，先用png顶替
\centering
\vspace{-.1in}
  \includegraphics[width=0.98\linewidth]{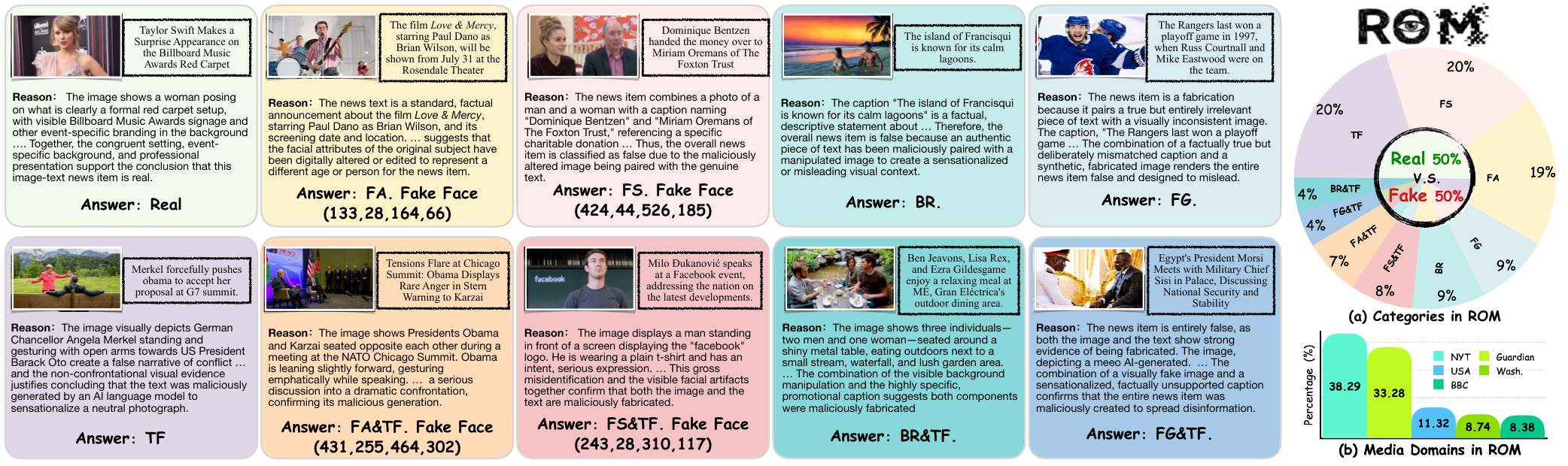}
  \vspace{-0.2cm}
  \caption {Overview of the ROM dataset. \textit{Left:} Representative samples spanning 9 manipulated and 1 real categories, ranging from face-centric edits to scene-level synthesis, each accompanied by a detailed reasoning annotation. \textit{Right:} Statistical distribution showing the diversity of manipulation types and the coverage of news media domains.}
  \vspace{-0.6cm}
  \label{fig:dataset_show}
\end{figure*}

To support our process reasoning-focused training as well as construct a comprehensive benchmark for generalization evaluation, we prepare Reasoning-enhanced analysis for Omnibus Manipulation (ROM) dataset, a  full-scene large-scale benchmark with thinking annotation. While incorporating face-centric samples from MDSM \citep{AMD_MDSM}, ROM significantly transcends previous boundaries by expanding the scope to scene-level synthesis and integrating logical reasoning. As illustrated in Fig.~\ref{fig:dataset_show}, the ROM dataset comprises \textbf{704,456} image-text pairs across five news domains and nine manipulation categories.

\noindent\textbf{Manipulation Scope Expansion.} \label{sec:data_expansion} ROM assimilates six face-specific classes from MDSM~\citep{AMD_MDSM}, comprising \textit{Original} (Orig), \textit{FaceAttribute} (FA), \textit{FaceSwap} (FS), \textit{TextFabrication} (TF), and their composites (FA\&TF, FS\&TF). Crucially, we extend this taxonomy by introducing four scene-level categories: \textit{BackgroundReplacement} (BR), \textit{FullGeneration} (FG), and their text-fabricated variants (BR\&TF, FG\&TF). This expansion moves beyond local facial edits to holistic scene synthesis, utilizing diverse generative models (e.g., SD3~\citep{SD3_ICML24}, FLUX.1~\citep{FLUX1_2025}) to ensure artifact diversity. 
% For a step-by-step breakdown of the sample generation process, please refer to the \hyperref[sec:ROM_pipeline]{Appendix E}.

\begin{figure}[t]
\centering
  \includegraphics[width=0.95\columnwidth]{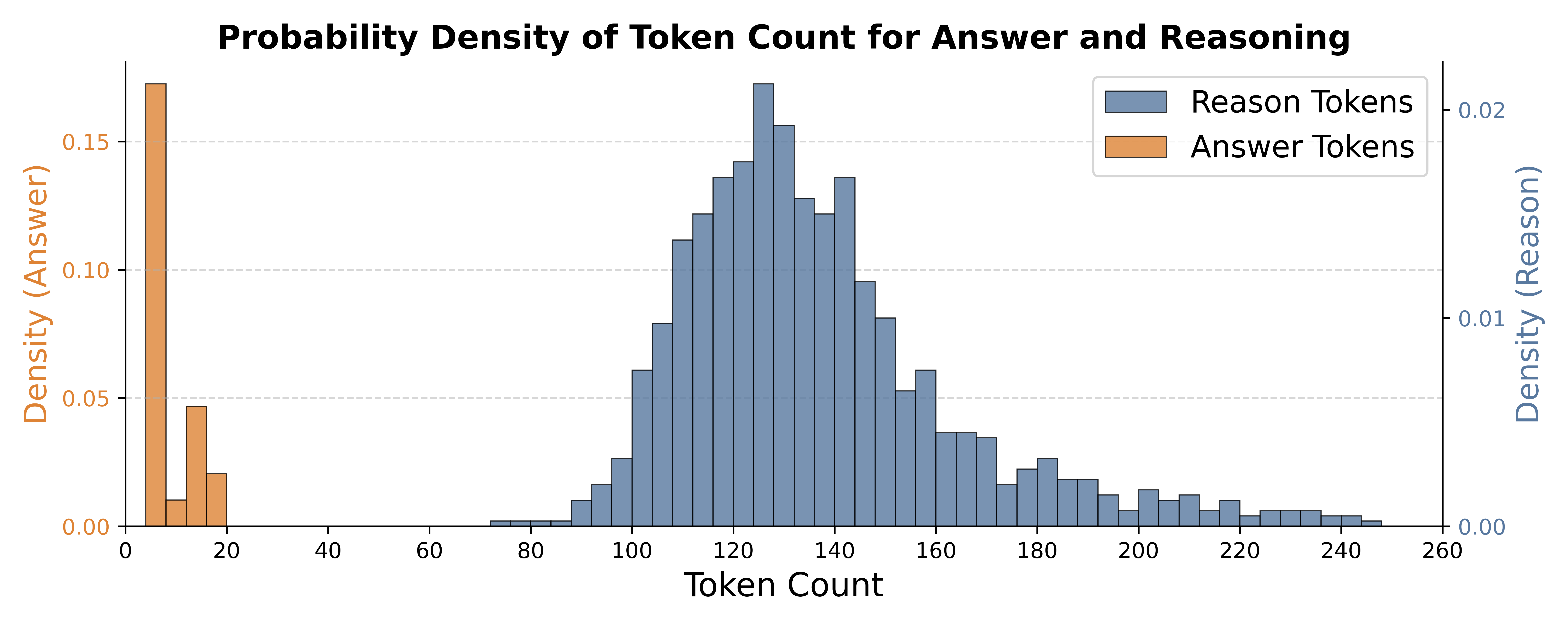}
    \vspace{-0.5cm}
  \caption {Probability Density of Token Count for Answer and Reasoning.}
  \vspace{-0.5cm}
  
  \label{fig:token_len_comp}
\end{figure}
\noindent\textbf{Reasoning Data Distillation.} We augment ROM with rationale annotations to enhance interpretability. Using InternVL3.5-30B \cite{wang2025internvl3}, we generate textual reasoning for each image-text pair given its manipulation label. Fig.~\ref{fig:dataset_show} visualizes these reasoning samples. As shown in Fig. \ref{fig:token_len_comp}, generated rationales typically peak around 130 tokens, offering significantly richer context than standard short answers ($<$20 tokens). 
% Detailed prompts are in the \hyperref[sec:supp_prompt]{Appendix G}.

\subsection{Architecture}
\begin{figure*}[t]

\centering
\vspace{-.1in}
  \includegraphics[width=\linewidth]{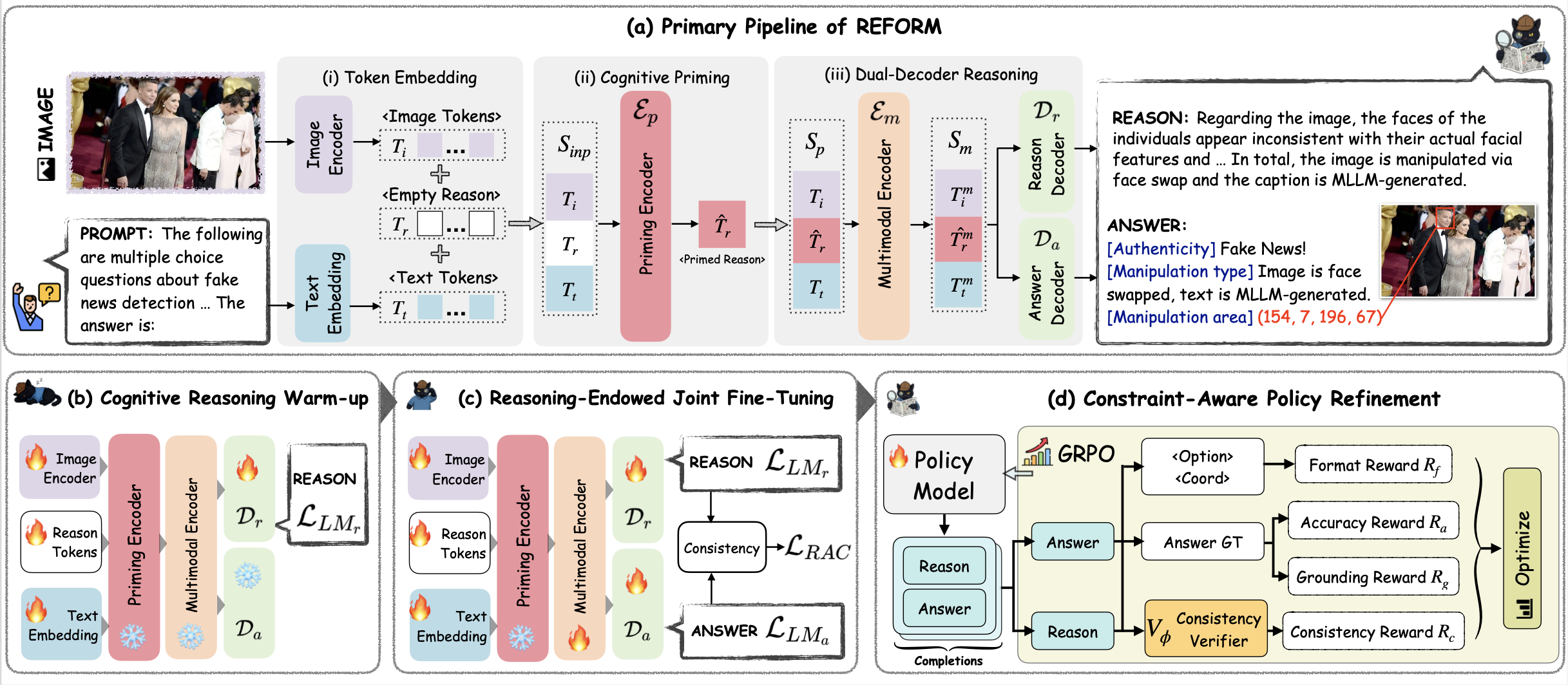}
  \vspace{-0.8cm}
  \caption{Overview of the REFORM framework and its three-stage training curriculum.
  (a) The primary pipeline employs a Cognitive Priming Encoder $\mathcal{E}_p$ and a Dual-Decoder structure, $\mathcal{D}_r$ and $\mathcal{D}_a$, for reasoning-driven detection.
  (b) Cognitive Reasoning Warm-up via partial freezing.
  (c) Reasoning-Endowed Joint Fine-Tuning incorporating the Reason-Answer Consistency Loss $\mathcal{L}_{RAC}$.
  (d) Constraint-Aware Policy Refinement using GRPO-based Reinforcement Learning to align forensic logic with the final verdict.}
\label{fig:model_arch}
\end{figure*}

Fig.~\ref{fig:model_arch}a illustrates the architecture of our proposed REFORM model, which ingests multimodal inputs and produces both detection and grounding results in textual form. 
REFORM is a sequence-to-sequence framework 
% that adopts DaViT \citep{ding2022davit} as its image encoder 
employs a novel encoder-dual-decoder structure as its backbone. We design a \emph{Cognitive Priming Module} to enhance the model's ability to capture forgery-related cues, and a \emph{Dual-Decoder} to strengthen its capability in interpreting forged evidence.

\noindent\textbf{Prompt Paradigm.}
The Prompt follows heuristic question(human)-answer(assistant) paradigm where the image-question pair serves as input and the text response constitutes REFORM’s output
% \footnote{Details and examples are given in \hyperref[sec:supp_prompt]{Appendix G}.}:

\begin{tcolorbox}[
    colback=gray!5,
    colframe=gray!40,
    title=\textbf{Prompt Template Definition},
    arc=2mm,
    fonttitle=\bfseries\small,
    boxrule=0.5pt,
    % --- 【极致压缩边距】 ---
    left=2mm, right=2mm, 
    top=1mm, bottom=1mm, % 上下留白极小
    middle=1mm,          % 分割线上下留白极小
    toptitle=1mm, bottomtitle=1mm, % 标题栏的留白
    % ----------------------------------
    fontupper=\footnotesize % 【使用更小的字体 footnotesize】
]
\label{prompt_for_REFORM}

    \noindent\texttt{\textbf{\#\#\#Human:}}\ptag{Task}\ptag{Options}\ptag{Grounding}
    \\[0.1em] % 稍微减小换行间距
    \noindent\texttt{\textbf{\#\#\#Assistant:}}\ptag{Response}\textcolor{bracketcolor}{[}\ptag{Coordinates}\textcolor{bracketcolor}{]}
    
    \tcbline 
    
    % --- 【使用 nosep 消除列表间距】 ---
    \begin{itemize}[nosep, leftmargin=*, labelsep=0.5em] 
        \item \ptag{Task}: Specifies manipulation detection objective \& pairs input.
        \item \ptag{Options}: Lists all candidate answers for the ROM task.
        \item \ptag{Grounding}: Triggers region localization conditionally.
        \item \ptag{Response}: Encapsulates the correct answers.
        \item \textcolor{bracketcolor}{[}\ptag{Coordinates}\textcolor{bracketcolor}{]}: Encloses edited region coordinates.
    \end{itemize}
\end{tcolorbox}

\noindent \textbf{Token Embedding.}
Through the image encoder, the input image is transformed into a sequence of \emph{<Image Tokens>}, denoted as $T_i$, where $n_v$ is the number of visual tokens and $d$ is the embedding dimension.  
Similarly, the question prompt is passed through an embedding layer to produce a sequence of \emph{<Text Tokens>}, represented as $T_t$.

\noindent \textbf{Cognitive Priming.}
This module is designed to perceive manipulation-related cues from the input data and distill these forensic signals into the Cognitive token sequence $T_r$. 
Structurally, the module is built upon the Cognitive Priming Encoder $\mathcal{E}_p$, instantiated as a standard Transformer encoder. 
Crucially, $\mathcal{E}_p$ operates in a parameter-frozen state; this constraint forces the learnable query tokens $T_r$ to actively extract and aggregate multimodal inconsistencies from the fixed semantic context provided by $T_i$ and $T_t$.
Formally, the tokens are concatenated to form ${S}_{inp}=[T_i; T_r; T_t]$, where $[\cdot; \cdot]$ denotes concatenation along the token dimension. The sequence ${S}_{inp}$ is then processed by $\mathcal{E}_p$:
\begin{equation}
\label{enocder_p}
     \mathcal{E}_p(S_{inp}) = [\hat{T}_i; \hat{T}_r; \hat{T}_t]. 
\end{equation}
We retain only $\hat{T}_r$ for further processing, which encapsulates the distilled manipulation signals.

\noindent \textbf{Dual-Decoder Reasoning.}
The primed reason tokens $\hat{T}_r$ are concatenated with the initial image tokens $T_i$ and text tokens $T_t$ to construct the composite input sequence $S_p = [T_i; \hat{T}_r; T_t]$. This sequence is encoded by the multimodal encoder $\mathcal{E}_m$ to yield the latent representation $S_m$. Subsequently, $S_m$ is fed into the Reason-Answer Dual-Decoder in parallel to generate textual responses. The Answer Decoder $\mathcal{D}_a$ generates a structured textual output comprising three distinct predictions, as shown in Fig.~\ref{fig:model_arch}a: (1) the authenticity classification (e.g., identifying the news as `fake'), (2) fine-grained manipulation types (e.g., detecting face-swapped and  AI-generated captions), and (3) manipulation localization (e.g., outputting coordinates to pinpoint the manipulated region). 
Meanwhile, the Reason Decoder $\mathcal{D}_r$ is tasked with producing a comprehensive explanation that rationalizes the model's verdict.

We adopt a dual-decoder architecture rather than a shared one for two reasons: (1) The expanded parameter space offers greater flexibility for optimizing both tasks independently, thereby reducing potential gradient conflicts during joint training. (2) It supports seamless switching between two modes: plugging reasoning decoder for reasoning mode, while un-equipping it for answer-only mode,  resulting in a more configurable framework.

%Notably, this decoupled design allows the reasoning branch to be bypassed during deployment, enabling rapid verdict generation without compromising accuracy.

% \zycnote{PS: Ablation: can try to unfreeze the Priming Encoder at this 1st stage and let the Priming Encoder inject semantic reason information into reason tokens}

\subsection{Cognitive Reasoning Warm-up}
\label{sec:stage1}
This initial phase is dedicated to aligning the model's cognitive process with the ground-truth forensic reasoning. Our primary objective is to specifically enhance the forensic reasoning capability by supervising the model with the distilled rationale annotations.

As illustrated in Fig.~\ref{fig:model_arch}b, we adopt a partial freezing strategy where the Multimodal Encoder $\mathcal{E}_m$, the Answer Decoder $\mathcal{D}_a$, and the Cognitive Priming Encoder $\mathcal{E}_p$ remain frozen. The optimization exclusively targets the learnable reason tokens $T_r$ and the Reason Decoder $\mathcal{D}_r$.
Under the supervision of the ground-truth rationales, the model is trained to reconstruct the detailed forensic explanation. And the reason token $T_r$ are forced to extract and encode specific manipulation traces that match the logical patterns in the reasoning annotations. 
We employ the standard causal language modeling objective as the reasoning generation loss, denoted as $\mathcal{L}_{LM_r}$, to guide this alignment:
\begin{equation}
    \mathcal{L}_{LM_r} = - \frac{1}{L} \sum_{t=1}^{L} \log P(y_t \mid y_{<t}, S_m),
    \label{eq:loss_reasoning}
\end{equation}
where $L$ is the length of the ground-truth rationale sequence, and $y_t$ denotes the $t$-th target token.

\subsection{Reasoning-Endowed Joint Fine-Tuning}
\label{sec:stage2}
Having established a coherent forensic logic foundation, we transition to the Supervised Fine-Tuning stage to endow the model's judgment logic with these forensic reasoning capabilities. This phase is executed through two key strategies:

\noindent \textbf{Dual-Stream Generative Optimization.} 
First, we activate the model's full capacity for simultaneous reasoning and judgment. As shown in Fig.~\ref{fig:model_arch}c, we unfreeze the entire backbone, including $\mathcal{E}_m$ and $D_a$. The model is now tasked with generating both the reasoning chain $R$ and the structured answer $A$. 
Thus, we optimize both streams via language modeling losses $\mathcal{L}_{LM_r}$ and $\mathcal{L}_{LM_a}$, where $\mathcal{L}_{LM_a}$ mirrors Eq.~\ref{eq:loss_reasoning} on answers annotations.

\noindent \textbf{Reason-Answer Semantic Alignment.} 
Solely minimizing generative losses independent of each other could lead to logical discrepancies, where the generated reasoning contradicts the final verdict. To bridge this potential semantic gap, we introduce the \textit{Reason-Answer Consistency Loss} ($\mathcal{L}_{RAC}$). This objective enforces a minimum semantic similarity between the representation of the rationale and the answer, ensuring the reasoning trajectory effectively substantiates the final answer.

% Let $\mathbf{H}^R$ and $\mathbf{H}^A$ \yxnote{not used!} denote the sequences of last hidden states from the reasoning decoder and the answer decoder, respectively. 
We derive fixed-size global embeddings $\mathbf{v}^R$ and $\mathbf{v}^A$ via mean pooling over hidden states, masked to exclude padding tokens to capture only valid semantic content. We then employ a margin-based hinge loss to penalize alignments falling below a threshold $\eta$:
\begin{equation}
    \mathcal{L}_{RAC} = \max \{ 0, \eta - \cos(\mathbf{v}^R, \mathbf{v}^A)\}.
\end{equation}
Overall, the full objectives for this Reasoning-Endowed Joint Fine-Tuning stage is:
\begin{equation}
    \mathcal{L}_{RJF} = \mathcal{L}_{LM_r} + \mathcal{L}_{LM_a} + \mathcal{L}_{RAC}.
\end{equation}

\subsection{Constraint-Aware Policy Refinement}
%\subsection{Reasoning Boost with Reinforce Refinement}
\label{sec:stage3}
To mitigate SFT's exposure bias and encourage the exploration of optimal reasoning, we adopt Group Relative Policy Optimization (GRPO) with a multi-dimensional reward function $\mathcal{R}$ consisting of four components:

\noindent \textbf{Consistency Reward ($\mathcal{R}_c$)} measures the semantic alignment between the generated reasoning chain $R$ and the predicted manipulation types, denoted as $\hat{c}_{i}$ (image-modal) and $\hat{c}_{t}$ (text-modal). We introduce a \textit{Consistency Verifier} $V_{\phi}$ to evaluate the logical entailment between the reasoning and these answer components.
We instantiate $V_{\phi}$ using a lightweight TinyBERT encoder~\citep{jiao-etal-2020-tinybert} equipped with two parallel classification heads for both modalities.
Prior to RL training, $V_{\phi}$ is pre-trained on ground-truth reason-label pairs, achieving over 99\% classification accuracy. This ensures $V_{\phi}$ can reliably deduce the correct manipulation category solely from a reasoning description.
The consistency reward is calculated by checking if $V_{\phi}$'s deductions match the predictions from $\mathcal{D}_a$:
\begin{equation}
    \mathcal{R}_c = \mathbb{I}(V_{\phi}^{i} = \hat{c}_{i}) + \mathbb{I}(V_{\phi}^{t} = \hat{c}_{t}),
\end{equation}
where $\mathbb{I}(\cdot)$ is the indicator function.

\noindent \textbf{Accuracy Reward ($\mathcal{R}_a$)} aligns predictions with ground-truth labels via binary verification ($\mathcal{R}_{bin}$) and fine-grained type recognition ($\mathcal{R}_{fin}$). Given ground truths $y_{bin}, c_{i}, c_{t}$ and predictions $\hat{y}, \hat{c}_{i}, \hat{c}_{t}$, we formulate the reward as:
\begin{align}
    \mathcal{R}_{bin} &= \mathbb{I}(\hat{y} = y_{bin}), \\
    \mathcal{R}_{fin} &= \mathbb{I}(\hat{c}_{i} = c_{i}) + \mathbb{I}(\hat{c}_{t} = c_{t}), \\
    \mathcal{R}_{a} &= \mathcal{R}_{bin} + \mathcal{R}_{fin}.
\end{align}

\noindent \textbf{Grounding Reward ($\mathcal{R}_g$)} evaluates the spatial precision by calculating the Intersection over Union (IoU) between the predicted bounding boxes $\hat{b}$ and the ground truth boxes $b$:
\begin{equation}
    \mathcal{R}_g = \text{IoU}(\hat{b}, b).
\end{equation}

\noindent \textbf{Format Reward ($\mathcal{R}_f$)} enforces strict adherence to the specified output structure, as defined in Sec.~\ref{prompt_for_REFORM}. Let $\hat{a}$ be the generated answer string and $\mathcal{S}$ be the set of valid regex patterns:
\begin{equation}
    \mathcal{R}_f = \mathbb{I}(\hat{a} \in \mathcal{S}).
\end{equation}

\noindent \textbf{Optimization Objective.} 
For each input prompt $x$, we sample a group of $G$ outputs $\{o_1, o_2, \dots, o_G\}$ from the current policy $\pi_\theta$. The total reward for the $i$-th output is the sum of the above components: $\mathcal{R}_{total}^{(i)} = \mathcal{R}_c + \mathcal{R}_a + \mathcal{R}_g + \mathcal{R}_f$.
To compute the advantage $A_i$, we normalize the rewards within the group to reduce variance. Let $r_i = \frac{\pi_\theta(o_i|x)}{\pi_{\text{old}}(o_i|x)}$ denote the probability ratio. The surrogate objective for the $i$-th sample is formulated as:
\begin{equation}
    \mathcal{J}_{\text{clip}}^{(i)} = \min \left( r_i A_i, \text{clip}\left( r_i, 1 \pm \epsilon \right) A_i \right).
\end{equation}
The final GRPO objective averages this surrogate objective while applying a KL-divergence penalty to prevent policy collapse:
\begin{equation}
    \mathcal{L}_{\text{G}} = \mathbb{E} \left[ \frac{1}{G} \sum_{i=1}^G \mathcal{J}_{\text{clip}}^{(i)} - \beta \mathbb{D}_{KL}(\pi_\theta || \pi_{\text{ref}}) \right].
\end{equation}

\section{Experiment}
% 关于所有实验模型的训练设置和验证指标的定义，请参考appendix
% For the implementation details of baseline models, the training settings, and the definitions of the evaluation metrics, please refer to the \hyperref[sec:Experimental_Setup]{Appendix}. 

Please refer to Appendix for implementation details and evaluation metric. 
Since rationale prediction is introduced solely to cultivate reasoning capabilities without ground-truth rationales, we validate its effectiveness implicitly through the performance of the primary detection and grounding tasks. 
% Qualitative examples are provided in \hyperref[sec:Case_Analysis]{Appendix F}.

%%%%<----MDSMv2对比试验大表(start)---->%%%%%%%
\definecolor{MintGray}{HTML}{F0F7F5}
\begin{table*}[t]
\caption{Comparison of multimodal learning methods on ROM, where the background \colorbox{lightgray!25}{gray} indicates the intra-domain test. The better results are in \textbf{bold}. AVG refers to the average performance across 5 news domains.}
\label{tab:baselines_Comparison_ROM}
\vspace{-0.3cm}
\centering
\begin{threeparttable}
\footnotesize
\setlength{\tabcolsep}{1mm}
\renewcommand{\arraystretch}{1.3} %%行间距宽一点，不然侧边竖排文字放不下
{\resizebox{\textwidth}{!}{\begin{tabular}{@{}l|ccccccccccccccccccc@{}}
\toprule[1.5pt]
\multirow{3}{*}{\rotatebox{90}{\textbf{Setting}}} & \multirow{3}*{\textbf{Method}} &\multicolumn{18}{c}{\textbf{Test Domain}} \\ 

\cline{3-20}
 &  & \multicolumn{3}{c}{NYT} & \multicolumn{3}{c}{Guardian} & \multicolumn{3}{c}{USA} & \multicolumn{3}{c}{Wash.} & \multicolumn{3}{c}{BBC} & \multicolumn{3}{c}{\textbf{AVG}}\\ [-0.05cm] 
%\cline{3-20}
\cmidrule(lr){3-5} \cmidrule(lr){6-8} \cmidrule(lr){9-11} \cmidrule(lr){12-14} \cmidrule(lr){15-17} \cmidrule(lr){18-20} \\[-0.35cm] 
 & & ACC & mAP & mIoU & ACC & mAP & mIoU & ACC & mAP & mIoU& ACC & mAP & mIoU & ACC & mAP & mIoU & ACC & mAP & mIoU \\[-0.05cm] 

 \midrule
 \multirow{7}{*}{\rotatebox{90}{Zero-Shot}}

 & Yi-VL-6B ~\citep{yiAL2024} & 31.03  & 21.86  & 2.67  & 20.13  & 19.18  & 1.14  & 19.65  & 12.76  & 1.72  & 26.15  & 25.47  & 1.79  & 29.67  & 17.38  & 1.11  & 25.33  & 19.33  & 1.69   \\

& DeepSeek-VL2-27B ~\citep{wu2024deepseekvl2}   & 45.74  & 34.01  & 8.21  & 39.76  & 27.67  & 7.13  & 31.78  & 36.91  & 7.21  & 29.25  & 29.03  & 6.90  & 33.01  & 29.62  & 4.03  & 35.91  & 31.45  & 6.70
\\

 & LLaVA-v1.6-34B ~\citep{LLAVA_NeurIPS}   & 47.66  & 41.09  & 7.45  & 38.54  & 31.12  & 6.76  & 32.89  & 37.27  & 6.43  & 31.09  & 31.23  & 6.78  & 36.00  & 29.58  & 4.12  & 37.24  & 34.06  & 6.31
\\

 & Qwen2.5-VL-72B ~\citep{bai2025qwen25}  & 50.74  & 42.24  & 12.79  & 40.18  & 32.70  & 11.08  & 33.64  & 37.60  & 10.99  & 35.11  & 34.29  & 10.28  & 37.11  & 30.51  & 7.69  & 39.35  & 35.47  & 10.56 \\

 % & Qwen3-VL-235B ~\citep{Qwen3_blog}    \\
 
 & GPT-4o ~\citep{hurst2024gpt4o}  & \textbf{51.98}  & 42.15  & \textbf{16.42}  & \textbf{42.18}  & 32.09  & \textbf{11.79}  & 46.08  & 38.31  & 14.36  & 39.18  & 33.41  & 11.07  & 37.21  & 30.72  & 10.47  & 43.33  & 35.34  & 12.82 \\
 
 & Gemini-2.5 ~\citep{gemini_25} & 50.86  & \textbf{42.73}  & 15.41  & 42.11  & \textbf{33.16}  & 11.54  & \textbf{49.42}  & \textbf{40.56}  & \textbf{15.24}  & \textbf{39.84}  & \textbf{38.70}  & \textbf{11.87}  & \textbf{39.54}  & \textbf{34.84}  & \textbf{11.71}  & \textbf{44.35}  & \textbf{38.00}  & \textbf{13.15}  \\

\cline{2-20}

  & MMD-Agent-34B ~\citep{liu2024mmfakebench}& 61.34  & 41.63  & 40.46  & 61.13  & 36.24  & 55.67  & 62.32  & 47.11  & 63.54  & 51.23  & 42.42  & 63.07  & 51.23  & 45.34  & 71.47  & 57.45  & 42.55  & 58.84
\\

 \midrule

\multirow{10}{*}{\rotatebox{90}{Train on NYT}}  & 
\multicolumn{19}{c}{\cellcolor{MintGray} Fine-tuned LVLMs}\\

& Qwen2.5-3B ~\citep{bai2025qwen25}   & \cellcolor{lightgray!25}86.76  & \cellcolor{lightgray!25}57.54  & \cellcolor{lightgray!25}65.34  & 80.04  & 41.34  & 62.78  & 72.07  & 34.82  & 62.59  & 74.87  & 38.43  & 55.75  & 73.13  & 39.98  & 70.00  & 77.38  & 42.42  & 63.29 \\

&LLaVa-v1.6-7B ~\citep{LLAVA_NeurIPS}   & \cellcolor{lightgray!25}94.42  & \cellcolor{lightgray!25}81.23  & \cellcolor{lightgray!25}83.87  & 83.45  & 60.37  & 58.06  & 70.16  & 53.96  & 61.36  & 69.80  & 53.13  & 59.27  & 84.59  & 52.26  & 73.99  & 80.48  & 60.19  & 67.31\\

&  \multicolumn{19}{c}{\cellcolor{MintGray} Deepfake Detectors}\\

& ViLT ~\citep{ViLT_ICML} & \cellcolor{lightgray!25}80.17  & \cellcolor{lightgray!25}69.71  & \cellcolor{lightgray!25}32.07  & 67.02  & 55.16  & 30.61  & 66.14  & 56.71  & 31.18  & 71.71  & 58.31  & 29.61  & 71.03  & 45.11  & 30.62  & 71.21  & 57.00  & 30.82  \\

 & HAMMER ~\citep{DGM4}   & \cellcolor{lightgray!25}81.49  & \cellcolor{lightgray!25}73.97  & \cellcolor{lightgray!25}59.53  & 67.94  & 56.24  & 43.29  & 68.05  & 61.35  & 49.12  & 70.94  & 57.61  & 49.02  & 73.63  & 54.50  & 45.05 & 72.41  & 60.73  & 49.20  \\
 
 & HAMMER++ ~\citep{DGM4_TPAMI}    & \cellcolor{lightgray!25}82.91  & \cellcolor{lightgray!25}74.07  & \cellcolor{lightgray!25}59.92  & 66.42  & 57.04  & 44.36  & 67.55  & 63.22  & 51.04  & 71.11  & 58.12  & 49.62  & 74.98  & 52.01  & 45.39 & 72.59  & 60.89  & 50.07 \\
 
 & FKA-Owl ~\citep{liu2024fkaowl}  & \cellcolor{lightgray!25}95.76  & \cellcolor{lightgray!25}89.66  & \cellcolor{lightgray!25}74.13  & 82.22  & 64.42  & 65.17  & \textbf{83.64}  & 63.33  & 73.74  & 78.37  & 59.84  & 64.52  & 85.83  & 61.03  & 69.13  & 85.17  & 67.66  & 69.34 \\

 & AMD ~\citep{AMD_MDSM} & \cellcolor{lightgray!25}94.13  & \cellcolor{lightgray!25}89.28  & \cellcolor{lightgray!25}\textbf{89.22}  & 85.52  & 68.98  & 75.17  & 81.32  & 67.72  & 73.74  & 80.14  & 65.01  & 74.15  & 88.48  & 64.68  & 74.64  & 85.92  & 71.13  & 77.38\\

 & \textbf{REFORM (Ours)}     & \cellcolor{lightgray!25}\textbf{96.69}  & \cellcolor{lightgray!25}\textbf{91.76}  & \cellcolor{lightgray!25}88.34  & \textbf{86.87}  & \textbf{73.18}  & \textbf{76.86}  & 83.35  & \textbf{69.16}  & \textbf{75.26}  & \textbf{83.87}  & \textbf{72.15}  & \textbf{76.40}  & \textbf{90.34}  & \textbf{74.17}  & \textbf{75.52} & \textbf{88.22}  & \textbf{76.08}  & \textbf{78.48} \\
 
\midrule

\multirow{10}{*}{\rotatebox{90}{Train on Guardian}} & 
\multicolumn{19}{c}{\cellcolor{MintGray} Fine-tuned LVLMs}\\

& Qwen2.5-3B ~\citep{bai2025qwen25}   & 61.90 &	22.23 &	61.99 &	\cellcolor{lightgray!25}92.76 &	\cellcolor{lightgray!25}75.62 &	\cellcolor{lightgray!25}77.23 &	67.59 &	44.62 &	63.44 &	70.76 &	36.04 &	67.64 &	76.21 &	49.17 &	72.38 &	73.84 	&45.54 &	68.54  \\

&LLaVa-v1.6-7B ~\citep{LLAVA_NeurIPS}  & 63.15 & 	23.14 &	65.35 &	\cellcolor{lightgray!25}93.66 &	\cellcolor{lightgray!25}75.13 	& \cellcolor{lightgray!25}83.37 &	67.68 &	45.87 &	66.72 &	69.08 &	33.74 &	62.54 &	80.63 &	54.56 &	77.52 &	74.84 	&46.49 	&71.10  \\

&  \multicolumn{19}{c}{\cellcolor{MintGray} Deepfake Detectors}\\

& ViLT ~\citep{ViLT_ICML} & 68.06 & 47.36 & 38.44 & \cellcolor{lightgray!25}88.50 & \cellcolor{lightgray!25}89.79 & \cellcolor{lightgray!25}60.36 & 65.37 & 51.17 & 46.13 & 63.30 & 58.34 & 47.02 & 78.89 & 49.15 & 58.69 & 72.82 & 59.16 & 50.13 \\

& HAMMER ~\citep{DGM4} & 71.15 & 49.74 & 47.24 & \cellcolor{lightgray!25}90.69 & \cellcolor{lightgray!25}92.50 & \cellcolor{lightgray!25}69.25 & 63.50 & 51.47 & 57.41 & 64.12 & 57.02 & 58.42 & 80.04 & 43.43 & 73.26 & 73.90 & 58.83 & 61.12 \\

& HAMMER++ ~\citep{DGM4_TPAMI} & 70.79 & 47.57 & 47.61 & \cellcolor{lightgray!25}89.63 & \cellcolor{lightgray!25}91.90 & \cellcolor{lightgray!25}71.67 & 61.53 & 50.85 & 58.39 & 62.58 & 55.17 & 59.66 & 81.14 & 43.91 & 73.35 & 73.13 & 57.88 & 62.14 \\

& FKA-Owl ~\citep{liu2024fkaowl} & \textbf{75.46} & 45.14 & 56.41 & \cellcolor{lightgray!25}91.83 & \cellcolor{lightgray!25}70.28 & \cellcolor{lightgray!25}72.51 & 70.32 & 30.65 & 76.85 & 70.06 & 31.56 & 72.34 & 83.14 & 42.93 & 63.16 & 78.16 & 44.11 & 68.25 \\

& AMD ~\citep{AMD_MDSM} & 74.23 & 42.98 & 64.49 & \cellcolor{lightgray!25}91.07 & \cellcolor{lightgray!25}76.54 & \cellcolor{lightgray!25}90.16 & 71.50 & 35.76 & 80.42 & 71.46 & 37.84 & 80.94 & \textbf{83.42} & 46.18 & 76.13 & 78.34 & 47.86 & 78.43 \\

& \textbf{REFORM (Ours)} & 74.38 & \textbf{63.62} & \textbf{67.53} & \cellcolor{lightgray!25}\textbf{94.04} & \cellcolor{lightgray!25}\textbf{92.90} & \cellcolor{lightgray!25}\textbf{94.07} & \textbf{79.16} & \textbf{58.48} & \textbf{84.56} & \textbf{78.04} & \textbf{59.49} & \textbf{83.41} & 81.98 & \textbf{64.26} & \textbf{77.75} & \textbf{81.52} & \textbf{67.75} & \textbf{81.46} \\
\bottomrule[1.5pt]
\end{tabular}}}
% \vspace{-0.1cm}
\end{threeparttable}

\end{table*}
%%%%<----MDSMv2对比试验大表(end)---->%%%%%%%

%%%%<----MMfakeBench试验大表(start) 双栏显示---->%%%%%%%
\begin{table*}[h!]
\centering
\caption{Comparison of zero-shot binary detection performance on the MMFakeBench validation and test sets. We compare our proposed REFORM with baseline models using ``Standard'' and ``MMD-Agent'' prompting paradigms as defined in the MMFakeBench paper. Baseline results are cited from the original MMFakeBench publication.}
\vspace{-.1in}
\label{tab:mmfakebench_binary}
\resizebox{\textwidth}{!}{% 自动调整表格宽度适应页面
\begin{tabular}{ccccccccccc}
\toprule
\textbf{Model} & \textbf{Language} & \textbf{Prompt} & \multicolumn{4}{c}{\textbf{Validation (1000)}} & \multicolumn{4}{c}{\textbf{Test (10000)}} \\
\cmidrule(lr){4-7} \cmidrule(lr){8-11}
\textbf{Name} & \textbf{Model} & \textbf{Method} & \textbf{F1} & \textbf{Precision} & \textbf{Recall} & \textbf{ACC} & \textbf{F1} & \textbf{Precision} & \textbf{Recall} & \textbf{ACC} \\
\midrule

% --- 7B Section ---
\rowcolor{MintGray} \multicolumn{11}{c}{\textbf{LVLMs with 7B Parameter}} \\
% Otter-Image & MPT-7B & Standard & 7.9 & 4.1 & 4.5 & 7.9 & 8.6 & 32.4 & 5.0 & 8.6 \\
% MiniGPT4 & Vicuna-7B & Standard & 40.4 & 38.2 & 45.7 & 63.1 & 41.7 & 41.0 & 47.4 & 65.2 \\
InstructBLIP~\citep{NEURIPS2023_InstructBLIP} & Vicuna-7B~\citep{NeurIPS_vicuna} & Standard & 14.7 & 30.8 & 13.2 & 8.1 & 16.1 & 40.5 & 14.2 & 8.8 \\
Qwen-VL~\citep{QwenVL} & Qwen-7B~\citep{QwenVL} & Standard & 43.6 & 50.6 & 44.9 & 60.3 & 44.0 & 51.6 & 45.2 & 60.5 \\
% VILA & LLaMA2-7B & Standard & 41.2 & 35.0 & 50.0 & 70.0 & 41.2 & 35.0 & 50.0 & 70.0 \\
PandaGPT~\citep{pandagpt_ACL23} & Vicuna-7B~\citep{NeurIPS_vicuna} & Standard & 24.6 & 60.6 & 50.5 & 30.9 & 24.1 & 61.7 & 50.4 & 30.6 \\
mPLUG-Owl2~\citep{ye2024mplugowl3longimagesequenceunderstanding} & LLaMA2-7B~\citep{Llama2_2023} & Standard & 47.2 & 64.9 & 52.3 & 70.6 & 48.7 & 71.1 & 53.3 & \textbf{71.4} \\
% BLIP2 & FlanT5-XL & Standard & 41.2 & 35.0 & 50.0 & 70.0 & 41.2 & 35.0 & 50.0 & 70.0 \\
LLaVA-1.6~\citep{LLAVA_NeurIPS} & Vicuna-7B~\citep{NeurIPS_vicuna} & Standard & 48.1 & 48.2 & 48.5 & 59.5 & 52.5 & 53.0 & 52.6 & 62.5 \\

% --- 13B Section ---
\rowcolor{MintGray} \multicolumn{11}{c}{\textbf{LVLMs with 13B Parameter}} \\

% \multirow{2}{*}{VILA} & \multirow{2}{*}{LLaMA2-13B} & Standard & 41.1 & 35.0 & 50.0 & 70.0 & 41.1 & 35.0 & 50.0 & 70.0 \\
% & & MMD-Agent & \textbf{56.5} & \textbf{62.2} & \textbf{56.9} & \textbf{70.3} & \textbf{56.6} & \textbf{64.3} & \textbf{57.2} & \textbf{71.2} \\

\multirow{2}{*}{InstructBLIP~\citep{NEURIPS2023_InstructBLIP}} & \multirow{2}{*}{Vicuna-13B~\citep{NeurIPS_vicuna}} & Standard & 41.1 & 35.0 & 49.9 & 69.9 & 41.1 & 35.0 & 49.9 & 69.8 \\
& & MMD-Agent & 51.3 & 53.4 & 54.0 & 53.1 & 47.9 & 50.1 & 50.1 & 49.9 \\

% \multirow{2}{*}{BLIP2} & \multirow{2}{*}{FlanT5-XXL} & Standard & 31.6 & \textbf{63.4} & 53.6 & 35.5 & 30.6 & \textbf{64.9} & 53.4 & 34.9 \\
% & & MMD-Agent & \textbf{51.5} & 53.4 & \textbf{54.0} & \textbf{53.6} & \textbf{51.8} & 54.0 & \textbf{54.7} & \textbf{53.5} \\

\multirow{2}{*}{LLaVA-1.6~\citep{LLAVA_NeurIPS}} & \multirow{2}{*}{Vicuna-13B~\citep{NeurIPS_vicuna}} & Standard & 41.1 & 35.0 & 50.0 & 69.7 & 42.3 & 57.3 & 50.1 & 69.5 \\
& & MMD-Agent & 51.8 & 66.7 & 54.6 & \textbf{71.4} & 50.2 & 67.3 & 53.9 & 71.3 \\

% --- REFORM ours ---
\rowcolor{MintGray} \multicolumn{11}{c}{\textbf{Ours}} \\
REFORM(Ours) & Florence2-0.3B~\citep{Florence2_CVPR} & Ours (Sec.\ref{prompt_for_REFORM}) & \textbf{74.9} & \textbf{74.5} & \textbf{75.4} & 64.7 & \textbf{74.1} & \textbf{74.1} & \textbf{74.0} & 63.7\\

\bottomrule
\end{tabular}
}
\vspace{-0.45cm}
\end{table*}
%%%%<----MMfakeBench试验大表(end)---->%%%%%%%

%%%%<----DGM4对比试验大表(start)---->%%%%%%%
\begin{table*}[t]
\caption{Comparison of multimodal learning methods on DGM4 (\%), where the guardian domain with background \colorbox{lightgray!25}{gray} is intra-domain. $\text{P}_{tok}$ is Precision of fake token grounding.}%The better results in each group are in boldface. AVG refers to the average performance across the four news domains.}
\label{tab:DGM4_baselines_Comparison}
\vspace{-.1in}
% \vspace{-0.3cm}
\centering
\begin{threeparttable}
\footnotesize
\setlength{\tabcolsep}{1mm}
\renewcommand{\arraystretch}{1.3} %%行间距宽一点，不然侧边竖排文字放不下
{\resizebox{\textwidth}{!}{\begin{tabular}{@{}ccccccccccccccccccccc@{}}
\toprule[1.5pt]
 \multirow{3}*{\textbf{Method}} &\multicolumn{20}{c}{\textbf{Test Domain}} \\ 

\cline{2-21}

 & \multicolumn{4}{c}{Guardian} & \multicolumn{4}{c}{USA} & \multicolumn{4}{c}{Wash.} & \multicolumn{4}{c}{BBC} & \multicolumn{4}{c}{\textbf{AVG}}\\ [-0.05cm] 
\cmidrule(lr){2-5} \cmidrule(lr){6-9} \cmidrule(lr){10-13} \cmidrule(lr){14-17} \cmidrule(lr){18-21}  \\[-0.35cm] 
 & ACC & mAP  & $\text{P}_{tok}$& mIoU & ACC  & mAP & $\text{P}_{tok}$& mIoU& ACC & mAP & $\text{P}_{tok}$ & mIoU& ACC & mAP  & $\text{P}_{tok}$& mIoU& ACC & mAP & $\text{P}_{tok}$& mIoU  \\[-0.05cm] 

 \midrule

\multicolumn{21}{c}{\cellcolor{MintGray} Fine-tuned LVLMs}\\

Qwen2.5-3B ~\citep{bai2025qwen25} & \cellcolor{lightgray!25}61.57  & \cellcolor{lightgray!25}37.36  & \cellcolor{lightgray!25}61.35  & \cellcolor{lightgray!25}40.19  & 60.65  & 34.25  & 70.19  & 33.28  & 51.79  & 32.23  & 63.23  & 35.11  & 62.20  & 38.02  & 61.22  & 41.23  & 59.05  & 35.47  & 64.00  & 37.45
\\

LLaVa-v1.6-7B ~\citep{LLAVA_NeurIPS} & \cellcolor{lightgray!25}68.67  & \cellcolor{lightgray!25}46.26  & \cellcolor{lightgray!25}65.71  & \cellcolor{lightgray!25}42.30  & 62.54  & 37.48  & 71.24  & 35.63  & 63.16  & 40.27  & 71.03  & 34.22  & 66.14  & 46.44  & 62.17  & 42.18  & 65.13  & 42.61  & 67.54  & 38.58 \\

\multicolumn{21}{c}{\cellcolor{MintGray} Deepfake Detectors}\\

ViLT ~\citep{ViLT_ICML} & \cellcolor{lightgray!25}68.27  & \cellcolor{lightgray!25}42.29  & \cellcolor{lightgray!25}69.87  & \cellcolor{lightgray!25}43.19  & 52.79  & 31.28  & 62.11  & 33.78  & 55.76  & 33.26  & 57.17  & 31.10  & 44.14  & 39.68  & 59.06  & 21.96  & 55.24  & 36.63  & 62.05  & 32.51 \\

HAMMER ~\citep{DGM4}   & \cellcolor{lightgray!25}78.34  & \cellcolor{lightgray!25}66.79  & \cellcolor{lightgray!25}78.27  & \cellcolor{lightgray!25}61.09  & 64.97  & 40.49  & 73.76  & 40.51  & 63.54  & 40.26  & 76.13  & 38.53  & 54.97  & 40.84  & 81.48  & 43.74  & 65.46  & 47.10  & 77.41  & 45.97 \\ 
        
HAMMER++ ~\citep{DGM4_TPAMI}  & \cellcolor{lightgray!25}79.13  & \cellcolor{lightgray!25}67.11  & \cellcolor{lightgray!25}78.24  & \cellcolor{lightgray!25}62.15  & 65.25  & 40.74  & 73.24  & 41.14  & 63.83  & 40.34  & 76.17  & 38.21  & 54.24  & 41.25  & 81.73  & 43.23  & 65.61  & 47.36  & 77.35  & 46.18  \\ 
        
FKA-Owl ~\citep{liu2024fkaowl}  & \cellcolor{lightgray!25}82.97  & \cellcolor{lightgray!25}53.86  & \cellcolor{lightgray!25}87.70  & \cellcolor{lightgray!25}65.69  & 67.57  & 38.97  & 79.44  & 32.57  & 67.05  & 37.70  & 81.55  & 31.86  & 70.26  & 40.20  & 84.54  & 46.48  & 71.96  & 42.68  & 83.31  & 44.15 \\ 
        
 AMD ~\citep{AMD_MDSM} & \cellcolor{lightgray!25}84.61  & \cellcolor{lightgray!25}68.50  & \cellcolor{lightgray!25}82.78  & \cellcolor{lightgray!25}81.24  & 70.62  & 43.20  & 75.73  & 41.99  & 70.28  & 43.36  & 77.76  & 39.05  & \textbf{72.37}  & 56.57  & 83.76  & 45.20  & 74.47  & 52.91  & 80.01  & 51.87 \\ 

\textbf{REFORM (ours)} & \cellcolor{lightgray!25}\textbf{91.10}  & \cellcolor{lightgray!25}\textbf{84.30}  & \cellcolor{lightgray!25}\textbf{89.92}  & \cellcolor{lightgray!25}\textbf{83.88}  & \textbf{71.95}  & \textbf{59.08}  & \textbf{81.23}  & \textbf{55.19}  & \textbf{72.84}  & \textbf{59.53}  & \textbf{83.13}  & \textbf{53.79}  & 70.70  & \textbf{59.97}  & \textbf{86.37}  & \textbf{47.92}  & \textbf{76.65}  & \textbf{65.72}  & \textbf{85.17}  & \textbf{60.19} \\

\bottomrule[1.5pt]
\end{tabular}}}
% \vspace{-1cm}
\end{threeparttable}
% \vspace{-0.5cm}
\end{table*}
%%%%<----DGM4对比试验大表(end)---->%%%%%%%

%%%%<----消融实验表组(start)---->%%%%%%%
\begin{table*}[t]
    \centering
    \caption{Ablation of components (a), reason token length (b), $\eta$ sensitivity (c), reward (d), and completion (e-f).}
    \label{tab:complex_layout}
    \vspace{-.1in}
    % ==========================
    % 第一行：3 个表格
    % ==========================
    
    % --- 表格 (a) [0.36\textwidth] ---
    \subfloat[Components Ablation.]{
        \begin{minipage}{0.36\textwidth}
            \centering
            % 请确保你的 preamble 中定义了 \tablestyle
            \tablestyle{1pt}{1.0}\tiny 
            \begin{tabular}{@{}cccccccccc@{}}
                \multicolumn{4}{c}{Components}
                  & \multicolumn{3}{c}{NYT}
                  & \multicolumn{3}{c}{Guardian} \\[-0.08cm]
                \cmidrule(lr){1-4} \cmidrule(lr){5-7} \cmidrule(lr){8-10} \\[-0.35cm]
                $\text{LM}_\text{a}$ & $\text{LM}_\text{r}$ & RAC & GRPO  & ACC & mAP & mIoU & ACC & mAP & mIoU \\[-0.08cm]
                \midrule[1.2pt]
                \ding{51} &           &           &           & 84.88 & 66.16 & 75.98 & 72.18 & 45.86     & 78.72  \\
                \ding{51} & \ding{51} &           &           & 87.76 & 73.01 & 77.68 & 74.74 & 53.65     & 79.59   \\
                \ding{51} & \ding{51} & \ding{51} &           & 87.84 & 73.25 & 78.00 & 75.71 & 54.11     & 79.58           \\
                \ding{51} & \ding{51} & \ding{51} & \ding{51} & \textbf{88.22} & \textbf{76.08} & \textbf{78.48} & \textbf{81.52}  & \textbf{67.75}     & \textbf{81.64}           \\
            \end{tabular}
        \end{minipage}
    }
    \hfill
    % --- 表格 (b) [0.30\textwidth] ---
    \subfloat[Reason Token Length.]{
        \begin{minipage}{0.30\textwidth}
            \centering
            \tablestyle{1pt}{1.0}\tiny
            \begin{tabular}{@{}lcccccc@{}}
                \multirow{2}{*}{Len.} & \multicolumn{3}{c}{NYT} & \multicolumn{3}{c}{Guardian} \\[-0.08cm]
                \cmidrule(lr){2-4} \cmidrule(lr){5-7} \\[-0.35cm]
                & ACC & mAP & mIoU & ACC & mAP & mIoU \\[-0.08cm]
                \midrule[1.2pt]
                16 & 87.49 & 75.10 & 76.83 & 80.78 & 67.26 & 80.38  \\
                32 & \textbf{88.22} & 76.08 & \textbf{78.48} & \textbf{81.52} & \textbf{67.75} & \textbf{81.46} \\
                64 & 88.16 & \textbf{76.17} & 78.28 & 81.39 & 67.52 & 81.41 \\
            \end{tabular}
        \end{minipage}
    }
    \hfill
    % --- 表格 (c) [0.30\textwidth] ---
    \subfloat[Sensitivity of $\eta$.]{
        \begin{minipage}{0.30\textwidth}
            \centering
            \tablestyle{1pt}{1.0}\tiny
            \begin{tabular}{@{}lcccccc@{}}
                \multirow{2}{*}{$\eta$} & \multicolumn{3}{c}{NYT} & \multicolumn{3}{c}{Guardian} \\[-0.08cm]
                \cmidrule(lr){2-4} \cmidrule(lr){5-7} \\[-0.35cm]
                & ACC & mAP & mIoU & ACC & mAP & mIoU \\[-0.08cm]
                \midrule[1.2pt]
                -0.1 &  88.12 & 75.50 & 78.03 & 81.11 & 67.11 & 81.18 \\
                0.0 & \textbf{88.22} &  \textbf{76.08} &    \textbf{78.48} & 81.52 &  \textbf{67.75} &    \textbf{81.46}  \\
                0.1 & 88.05 &   75.87 & 78.08 &\textbf{81.94} &   67.67 & 81.19  \\
            \end{tabular}
        \end{minipage}
    }

    % ==========================
    % 换行区
    % ==========================
    \vspace{0.3cm} 
    
    % ==========================
    % 第二行：3 张图片
    % ==========================

    \subfloat[GRPO Reward Configurations.]{
        \begin{minipage}{0.36\textwidth} 
            \centering
            \vspace{-.2in}
            \includegraphics[width=\linewidth, height=3cm]{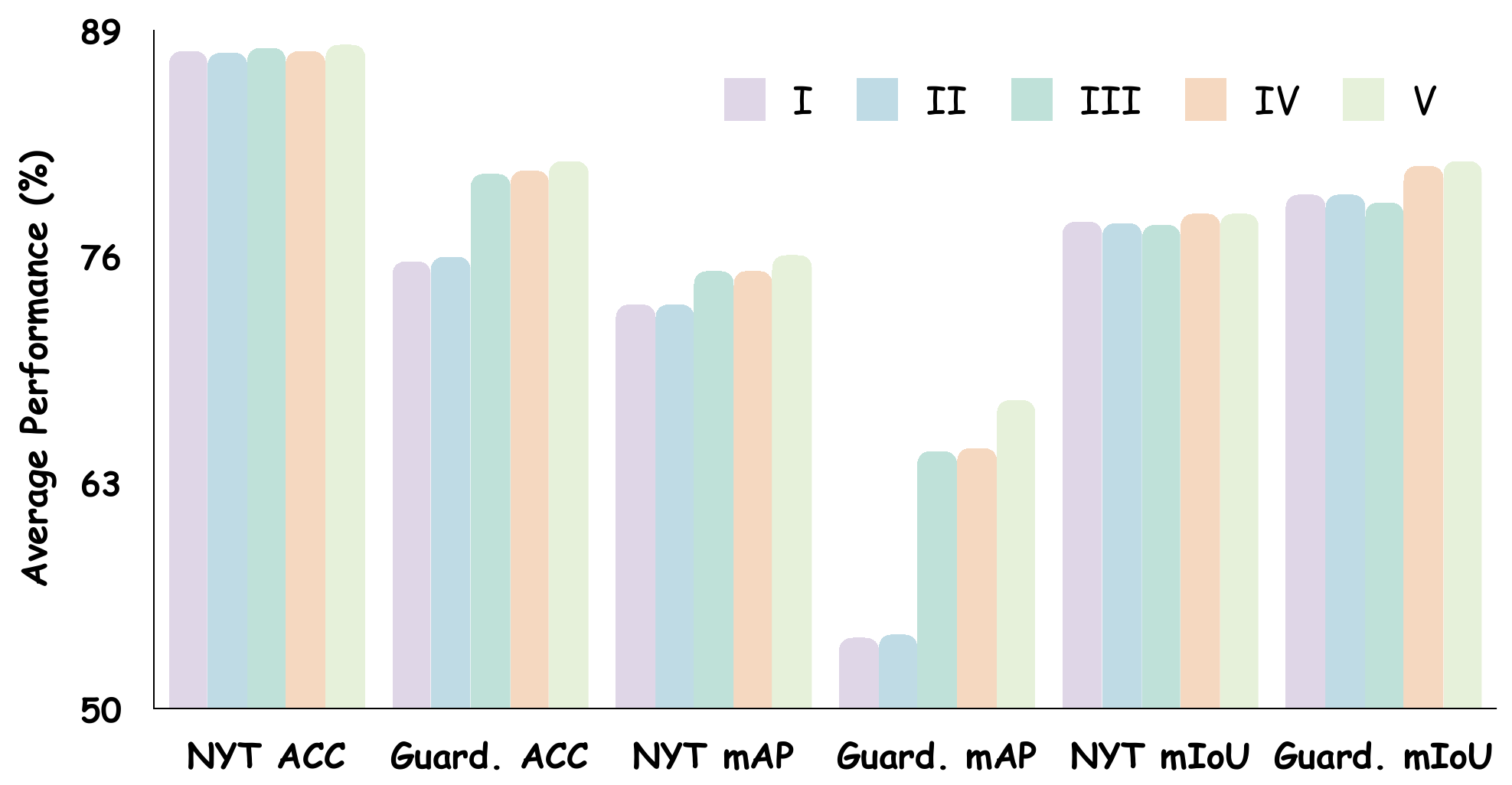}
        \end{minipage}
    }
    \hfill
    \subfloat[GRPO training on NYT.]{
        \begin{minipage}{0.30\textwidth} 
            \centering
            \vspace{-.2in}
            \includegraphics[width=\linewidth, height=3cm]{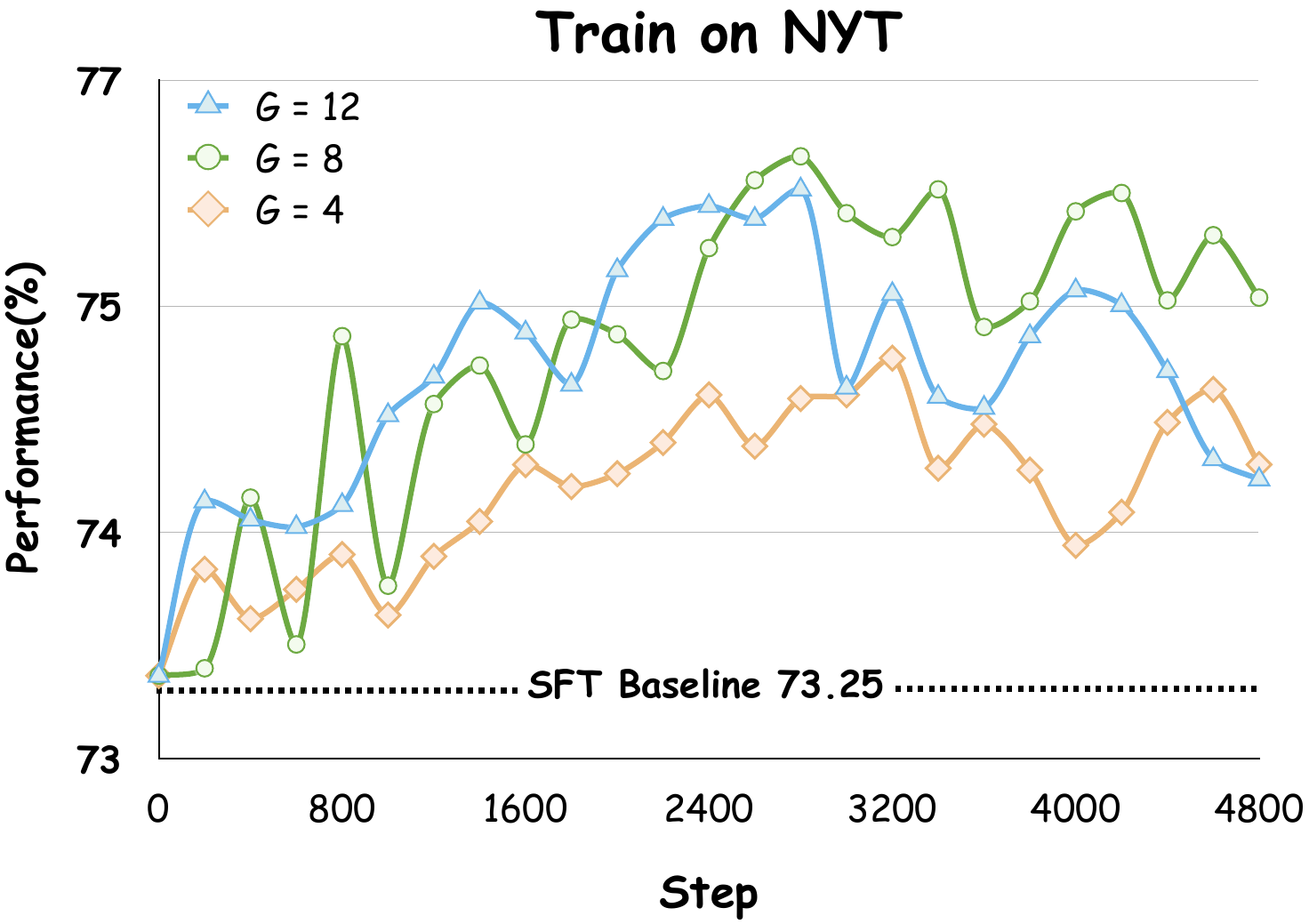}
        \end{minipage}
    }
    \hfill
    \subfloat[GRPO training on Guardian.]{
        \begin{minipage}{0.30\textwidth} 
            \centering
            \vspace{-.2in}
            \includegraphics[width=\linewidth, height=3cm]{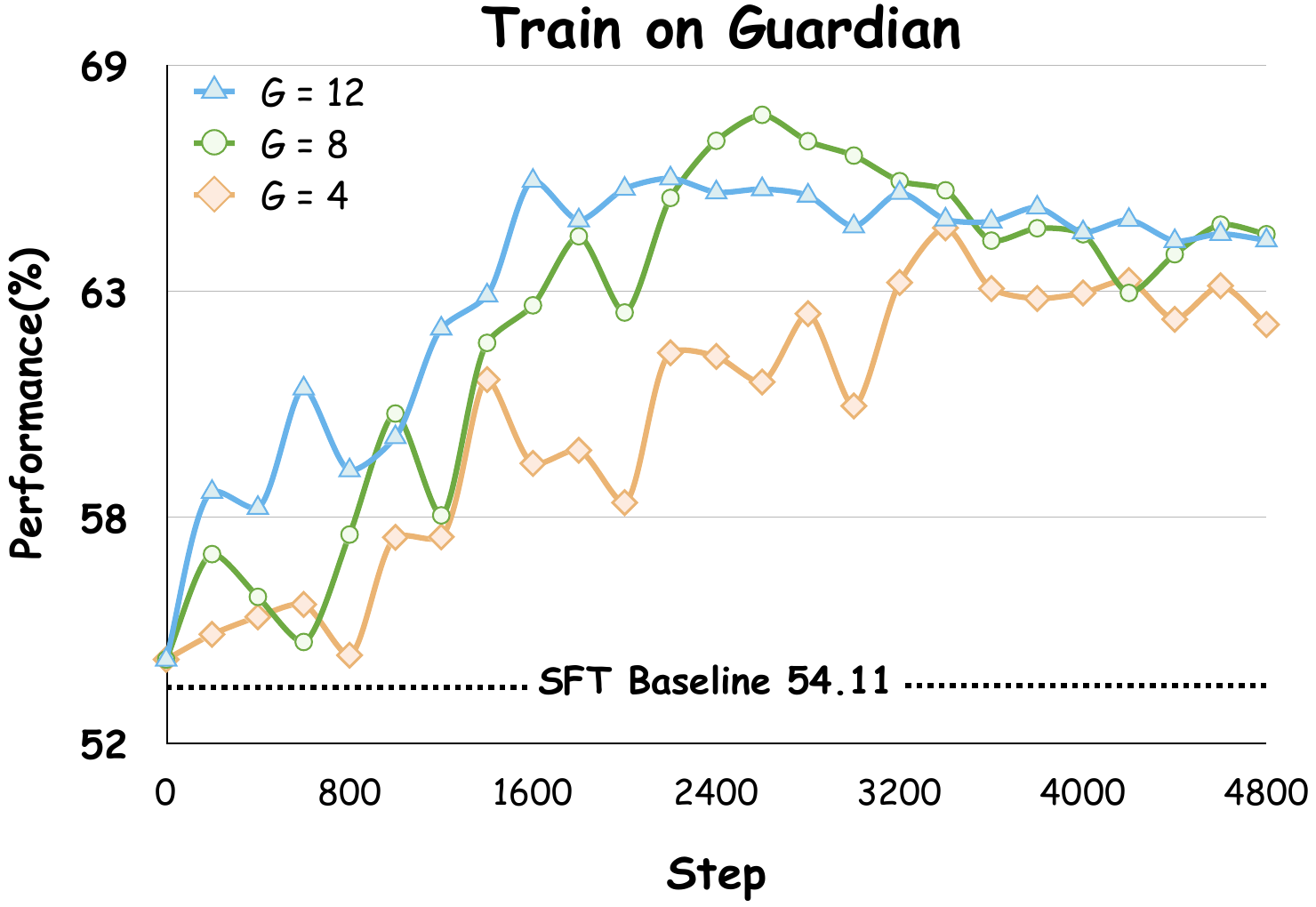}
        \end{minipage}
    }

    \vspace{-0.75cm} 
\end{table*}
%%%%<----消融实验表组(end)---->%%%%%%%

%%%%<----模型参数表(start)---->%%%%%%%
\begin{table}[t]
  \centering
  \caption{Efficiency comparison. We report total parameters and throughput (pairs/sec) on RTX 4090. Fast Mode refers to REFORM generating only prediction labels without the reasoning chain. %, enabling high-speed screening.
  }
  \vspace{-0.3cm}
  \label{tab:Efficiency_Comparison}
  % \scriptsize % resizebox 会自动调整大小，这里去掉 scriptsize 以免双重缩小
  \setlength{\tabcolsep}{1.5mm} % 适当的列间距
  \renewcommand{\arraystretch}{1.1} % 稍微增加行高，提升可读性
  \resizebox{\linewidth}{!}{% 使用 resizebox 自适应宽度
  \begin{tabular}{@{}lcccc@{}}
    \toprule
    \multirow{2}{*}{Method} 
    & \multicolumn{2}{c}{Params (M) $\downarrow$} 
    & \multicolumn{2}{c}{Throughput (p/s) $\uparrow$} \\[-0.05cm]
    \cmidrule(lr){2-3} \cmidrule(lr){4-5}
    & Total & Trainable & Train & Inference \\
    \midrule
    ViLT & 121.07 & 121.07 & 1.85 & 2.38 \\
    HAMMER(++) & 441.12 & 228.25 & 28.97 & 61.28 \\
    FKA-Owl & 6771.98 & 33.55 & 1.25 & 1.33 \\
    MMD-Agent & 34751.17 & 34447.66 & - & 0.02 \\ 
    AMD
      & 276.95
      & 276.95 
      & 5.55
      & 13.38 \\
    \midrule
    \rowcolor{gray!20}
     &  &  &  & \textit{Explainable}: 1.03 \\
    \rowcolor{gray!20}
    \multirow{-2}{*}{\textbf{REFORM (Ours)}} & \multirow{-2}{*}{\textbf{376.23}} & \multirow{-2}{*}{\textbf{376.23}} & \multirow{-2}{*}{\textbf{4.68}} & \textbf{\textit{Fast Mode}: 13.17} \\
    \bottomrule
  \end{tabular}%
  }
  \vspace{-0.5cm}
\end{table}
%%%%<----模型参数表(end)---->%%%%%%%

\subsection{Quantitative Results}
We evaluate REFORM on three comprehensive benchmarks: ROM, MMFakeBench~\citep{liu2024mmfakebench}, and DGM4~\citep{DGM4_TPAMI}.  \textbf{(1) Cross-Domain Generalization (Tab.~\ref{tab:baselines_Comparison_ROM}):} On ROM, REFORM significantly outperforms state-of-the-art (SOTA) baselines. In the challenging cross-domain setting (Train on NYT), REFORM achieves an average accuracy of 88.22, surpassing AMD (85.92) and HAMMER (72.41). Notably, REFORM also outperforms MMD-Agent-34B (57.45), an agentic pipeline that utilizes iterative reasoning and external knowledge (Wikipedia) retrieval.
\textbf{(2) Zero-shot Generalization (Tab.~\ref{tab:mmfakebench_binary}):}
Despite dealing with unseen manipulation types (e.g., manual PS editing) in MMFakeBench, REFORM achieves a remarkable F1 score of 74.9. It significantly outperforms 7B and 13B parameter LVLMs, demonstrating that forensic reasoning equips small-scale models (0.3B) with superior zero-shot generalizability compared to their larger counterparts.
% \textbf{(3) Superiority over Specialized Detectors (Tab.~\ref{tab:DGM4_baselines_Comparison}):} On the face-centric task DGM4, REFORM achieves SOTA performance, consistently outperforming specialized methods like FKA-Owl and AMD. While general LVLMs struggle with domain-specific artifacts ($<67$ AVG ACC), REFORM attains an impressive performance 76.65. This proves that optimizing forensic reasoning captures intrinsic manipulation traces more effectively than methods relying on external knowledge or invariant features.
\textbf{(3) Superiority over Specialized Detectors (Tab.~\ref{tab:DGM4_baselines_Comparison}):} 
REFORM establishes a new SOTA on the face-centric DGM4 benchmark. 
As observed, fine-tuned LVLMs struggle to detect subtle artifacts, yielding unsatisfactory mAP ($<47$). 
While specialized detectors like AMD and FKA-Owl, REFORM significantly outperforms them with average mAP 65.72. It shows that optimizing forensic reasoning captures intrinsic manipulation traces with greater cross-domain universality compared to feature-engineering approaches, leading to superior generalization.

% \yxnote{It should be noted that we do not evaluate the generated reasons, as rationale prediction is introduced solely to cultivate the model’s reasoning ability for manipulation detection and grounding. The rationales for explanation are neither the primary objective of this work nor accompanied by ground-truth annotations for reliable evaluation.}

\subsection{Ablation Studies}
We conduct extensive ablation studies, as summarized in Tab.~\ref{tab:complex_layout}. Note that all data presented in this table denote the cross-domain average performance.
\noindent\textbf{Impact of Component Modules.}
Tab.~\ref{tab:complex_layout}a validates our three-stage curriculum. The base model ($LM_a$) achieves 84.88 accuracy on NYT. Introducing the reasoning objective ($LM_r$) improves performance to 87.76, confirming that rationalization aids detection. The addition of consistency loss ($RAC$) and GRPO further boosts accuracy to 88.22 and mIoU to 78.48, demonstrating that enforcing logical consistency between reasoning and judgment is crucial for optimal performance.

\noindent\textbf{Impact of Reason Token Length.} 
We examine the effect of reason token length on cross-domain average performance. in Tab.~\ref{tab:complex_layout}b. Performance initially improves with length as the model captures more forensic details, peaking at 32 tokens with optimal ACC 88.22.

\noindent\textbf{Sensitivity Analysis of Consistency Margin.} 
As shown in Tab.~\ref{tab:complex_layout}c on $\eta$ in $\mathcal{L}_{RAC}$. We observe that the model exhibits robustness within a reasonable range, and $\eta=0.0$ yields the best trade-off, effectively penalizing semantic misalignment without disrupting the learning of distinct modal features.

\noindent\textbf{Impact of GRPO Reward Configurations.}
As illustrated in Tab.~\ref{tab:complex_layout}d, we analyze the cumulative effect of reward components. \RNum{1} denotes the baseline model without RL. Subsequent configurations sequentially introduce the Format Reward ($\mathcal{R}_f$, \RNum{2}), Accuracy Reward ($\mathcal{R}_a$, \RNum{3}), Grounding Reward ($\mathcal{R}_g$, \RNum{4}), and Consistency Reward ($\mathcal{R}_c$, \RNum{5}).
Observing the Guard. ACC, NYT mAP, and Guard. mAP metrics, a substantial performance leap occurs at \RNum{3} with the introduction of $\mathcal{R}_a$. Similarly, the integration of $\mathcal{R}_c$ in \RNum{5} yields another distinct performance boost. Regarding NYT ACC, since the baseline performance is already saturated (87.84), the scope for RL-driven improvement is naturally limited; thus, $\mathcal{R}_a$ and $\mathcal{R}_c$ yield only marginal gains, which is expected. Meanwhile, the inclusion of $\mathcal{R}_g$ significantly enhances mIoU across both domains.

% Overall, removing any reward component leads to suboptimal convergence. Specifically, the $\mathcal{R}_c$ proves most critical; without it, the policy fails to align the reasoning chain with the final verdict, resulting in a significant \yxnote{seems not significant from the figure} \zycnote{这一段是之前用来占位的描述，忘记注释掉了}drop in detection accuracy. 

\noindent\textbf{Impact of RL Training Configurations.}
We analyze the impact of generation group size $G$ throughout the training process in Tab.~\ref{tab:complex_layout}e-f, with performance measured by mAP. All RL configurations significantly outperform the SFT baselines (73.25 on NYT and 54.11 on Guardian), strongly validating our policy refinement. 
Notably, $G=8$ delivers the most robust results, achieving peak performance at $\sim$2.8k steps on NYT and $\sim$2.6k steps on Guardian. This configuration effectively balances exploration against stability. Visual analysis reveals that while larger groups ($G=12$) facilitate rapid initial convergence, they suffer from training instability and performance fluctuations in later stages. Conversely, smaller groups ($G=4$) restrict the exploration space, leading to suboptimal convergence and consistently lower accuracy.

\subsection{Discussion}
% \noindent\textbf{Efficiency Discussion.} 
% As shown in Tab.~\ref{tab:Efficiency_Comparison}, with only 376M parameters, REFORM is significantly more compact than FKA-Owl (6.7B) and MMD-Agent (34B). 
% Crucially, our dual-decoder design enables a Fast Mode (13.17 p/s) that bypasses the reasoning branch to achieve real-time screening speeds comparable to AMD. 
% Since the reasoning and answer branches operate independently and in parallel, Fast Mode incurs zero accuracy loss compared to the Explainable Mode.

\noindent\textbf{Efficiency Discussion.} 
As shown in Tab.~\ref{tab:Efficiency_Comparison}, with only 376M parameters, REFORM is significantly more compact than FKA-Owl (6.7B) and MMD-Agent (34B). 
Crucially, our dual-decoder design enables a Fast Mode (13.17 p/s) that bypasses the reasoning branch to achieve real-time screening speeds comparable to AMD. 
Since the reasoning and answer branches operate independently and in parallel, Fast Mode incurs zero accuracy loss compared to the Explainable Mode, because the answer decoder does not depend on the generated rationale at inference time. 
Thus, Fast Mode only removes the overhead of rationale generation, while leaving the final prediction unchanged.

% \textbf{Please refer to the \hyperref[sec:appendix_start]{Appendix} for more generalization results.}

\noindent\textbf{Faithfulness of Teacher Rationales.} 
To assess whether the distilled rationales are grounded in concrete multimodal evidence rather than merely restating manipulation labels, we conduct a blinded human audit on 350 ROM samples. As reported in Table~\ref{tab:human_audit_rom}, the rationales recover 83.7\% of the ground-truth visual evidence and 82.2\% of the ground-truth textual evidence. These results indicate that the teacher rationales are generally evidence-grounded rather than template-like label paraphrases. Although there is still room for improvement in rationale faithfulness, REFORM already yields substantial performance gains, supporting the effectiveness of our reasoning-driven learning paradigm.

\paragraph{Robustness to Teacher Quality.}
To examine whether REFORM's gains depend on a high-capacity annotator, we replace the original InternVL3.5-30B~\cite{wang2025internvl3} teacher with a much smaller Qwen2.5-VL-3B~\cite{bai2025qwen25} model on the Guardian setting of ROM. As shown in Table~\ref{tab:teacher_robustness_rom}, performance drops remain small, with decreases of only 0.84 in ACC, 1.46 in mAP, and 0.33 in mIoU. These results suggest that the effectiveness of REFORM is not solely driven by a powerful teacher model, but is largely preserved even when the reasoning supervision is distilled from a smaller annotator.

\begin{table*}[t]
\centering
\caption{Analysis on rationale faithfulness (a) and teacher robustness (b).}
\vspace{-0.2cm}

\small
\subfloat[Blinded human audit on ROM.\label{tab:human_audit_rom}]{
\begin{minipage}[t]{0.33\textwidth}
\centering
\begin{tabular}{lc}
\toprule
Metric & Score (\%) \\
\midrule
GT visual evidence recall  & 83.7 \\
GT textual evidence recall & 82.2 \\
\bottomrule
\end{tabular}
\end{minipage}
}
\hfill
\subfloat[Robustness to teacher quality on Guardian setting of ROM.\label{tab:teacher_robustness_rom}]{
\begin{minipage}[t]{0.63\textwidth}
\centering
\resizebox{\linewidth}{!}{
\begin{tabular}{lccc}
\toprule
Model & ACC & mAP & mIoU \\
\midrule
REFORM (InternVL3.5-30B) & \textbf{81.52} & \textbf{67.75} & \textbf{81.46} \\
REFORM (Qwen2.5-VL-3B) & 80.68 (-0.84) & 66.29 (-1.46) & 81.13 (-0.33) \\
\bottomrule
\end{tabular}
}
\end{minipage}
}

\vspace{-0.9cm}
\label{tab:additional_analysis}
\end{table*}
\section{Conclusion}
In this paper, we address the generalization challenge by shifting from result-oriented supervision to explicit forensic reasoning. We propose REFORM, which integrates cognitive priming and a dual-decoder architecture, optimized via a GRPO-based curriculum to align judgment with logical evidence. Additionally, we contribute ROM, a large-scale benchmark expanding the scope to scene-level synthesis with reasoning annotations. Experiments demonstrate that REFORM significantly outperforms SOTA methods in cross-domain and zero-shot settings, establishing a robust paradigm for interpretable forensics.
% \section{Limitations}
% \label{sec:limit}
% Despite the superior performance of REFORM, it has several limitations that merit future study. 

% \noindent \textbf{Dependence on Distilled Rationales.} Our training relies on reasoning annotations distilled from larger MLLMs (InternVL). Although we employ Reinforcement Learning to refine the policy, the model's upper bound is partially constrained by the quality and correctness of the teacher model's forensic knowledge. Hallucinations in the teacher model could potentially introduce noise into the reasoning supervision.

% \noindent \textbf{Inference Latency in Reasoning Mode.} While our Fast Mode achieves high throughput suitable for real-time screening, the Explainable Mode requires auto-regressive generation of reasoning chains, which increases computational cost ($1.03$ pairs/sec). Future work could explore non-autoregressive generation or token pruning to accelerate the reasoning process.

\section{Limitations}
\label{sec:limit}
Despite the superior performance of REFORM, it still has several limitations that merit future study. 

\noindent \textbf{Residual Dependence on Distilled Rationales.} 
REFORM relies on distilled rationales during training. Although our analyses show that these rationales are generally evidence-grounded and that REFORM is reasonably robust to teacher quality, the final performance can still be affected by the faithfulness of the distilled supervision. Since we do not explicitly optimize rationale quality in this work, further improving it may lead to stronger forensic reasoning and better overall performance.

\noindent \textbf{Inference Latency in Reasoning Mode.} 
While our Fast Mode is suitable for real-time screening, the Explainable Mode requires auto-regressive rationale generation, which increases computational cost ($1.03$ pairs/sec). Future work could explore more efficient rationale generation strategies.

% \noindent \textbf{Scope of Manipulation Types.} While ROM expands the manipulation scope significantly, the current synthesis pipeline primarily focuses on visual editing and text fabrication. More complex manipulations, such as temporal inconsistencies in video or subtle logical conflicts in long-form news reports, remain to be explored.
\section{Ethical Considerations}

\label{sec:ethics_statement}
This work adheres to the ACL Code of Ethics. The ROM dataset and associated analyses were created solely to support research on detecting and grounding multimodal manipulations. We recognize that assembling realistic synthetic examples entails dual-use risks: the same materials and procedures could be misused to produce deceptive content. To minimize harm, we adopt a harm-minimizing, controlled-release approach: we will not publish the generation pipeline, detailed prompts, or prompt--response pairs to prevent their exploitation by adversaries for generating harmful content; public distribution is limited to vetted, research-only access under a signed Data Usage Agreement (DUA); distributed images will carry conspicuous visual watermarks and standardized metadata tags; high-fidelity originals and sensitive metadata will be withheld; images of minors and clearly sensitive contemporary conflict content have been excluded; and we reserve the right to revoke access on evidence of misuse. Full technical and procedural details of these safeguards are documented in the dataset README file.

% \subsection{Large Language Model Usage Statement}
% This paper used Large Language Models solely for text polishing and expression refinement. No large language models were involved in other aspects of the research, including data collection, experimental design, result analysis, or conclusion derivation.

\section*{Acknowledgment}
This work was supported by the National Natural Science Foundation of China (NSFC) under Grant No. 62302140, and the National Key Research and Development Program of China under Grant No. 2023YFC3321600. The authors also gratefully acknowledge the support from the Guangdong Basic and Applied Basic Research Foundation (2025A1515012281), the Nanjing Municipal Science and Technology Bureau (202401035), and the University of Macau (MYRG-GRG2024-00077-FST-UMDF).

\bibliography{custom}
\clearpage

\appendix

% =================================================
% Appendix Start
% =================================================
\twocolumn[
  % --- 这里添加锚点 ---
  \phantomsection 
  \label{sec:appendix_start} 
  % --------------------
  \begin{@twocolumnfalse} 
    \centering
    
    {
      \Large \textbf{
    \raisebox{-0.15cm}{\includegraphics[height=1.3em]{AMDv2_logo.png}}~ Cultivating Forensic Reasoning for Generalizable Multimodal Manipulation Detection
      } \par
    }

    \vspace{0.5em}
    
    {\large \textit{
    Supplementary Materials} \par
    }
    \vspace{1em}

\vspace{1em}

\end{@twocolumnfalse}
]

\section{Experimental Setup}
\label{sec:Experimental_Setup}
\noindent \textbf{Implementation Details.}
All experiments are implemented in PyTorch and conducted on NVIDIA GeForce RTX 4090 GPUs using Distributed Data Parallel (DDP). 
The image encoder $\mathcal{E}_v$ adopts the DaViT architecture~\citep{ding2022davit}. 
Both the Multimodal Encoder $\mathcal{E}_m$ and the Cognitive Priming Encoder $\mathcal{E}_p$ are initialized with the pre-trained weights from Florence-2-B~\citep{Florence2_CVPR}. 
Similarly, the Reason Decoder $\mathcal{D}_r$ and Answer Decoder $\mathcal{D}_a$ are initialized using the Florence-2-B decoder weights. 
The semantic alignment verifier $V_{\phi}$ is built upon the TinyBERT-4L-312D architecture~\citep{jiao-etal-2020-tinybert}, comprising 4 transformer layers with a hidden dimension of 312. It employs a dual-head design—targeting 5-class image and 2-class text classification—where each head consists of a dense projection layer, GELU activation, and a dropout rate of 0.2. The input sequence length for $V_{\phi}$ is truncated to 320 tokens.

\noindent \textbf{Parameter Settings.}
Input images are resized to $224 \times 224$ and augmented with random horizontal flipping. 
For the supervised training phases (Stage 1 and Stage 2), we set the per-GPU batch size to 5 and train the model for 4 and 12 epochs, respectively. 
Optimization is performed using AdamW~\citep{AdamW_ICLR} with an initial learning rate of $1 \times 10^{-7}$ and a weight decay of 0.01. 
We employ a cosine learning rate scheduler with a linear warm-up: the learning rate increases to $1 \times 10^{-6}$ over the first 1,000 steps and subsequently decays to $1 \times 10^{-7}$. 
In Stage 3, initialized with the best Stage 2 checkpoint, we set the generation group size to $G=8$ to enhance exploration and optimization stability. The model is trained with a global batch size of 32 and a maximum sequence length of 1024. We utilize BF16 precision and Flash Attention 2~\citep{FlashAttention2} for efficiency.

\subsection{Baseline Settings}
We adapt six state-of-the-art multi-modal methods to the ROM setting for comparison. These methods encompass two conventional multi-modal manipulation detection models, two Multimodal Large Language Model (MLLM) -based detection frameworks, one MLLM-based agentic manipulation detection framework, and one general multi-modal learning approach:

\begin{itemize}
    % --- Conventional Models ---
    \item \textbf{HAMMER}~\citep{DGM4} is a pioneering model for multi-modal manipulation detection and grounding. It employs two unimodal encoders to extract visual and textual forgery features, which are then aligned through contrastive learning. Following this, a multi-branch transformer architecture with two specialized decoders is utilized for manipulation detection and grounding.
    
    \item \textbf{HAMMER++}~\citep{DGM4_TPAMI} is a more powerful model that builds upon HAMMER by integrating contrastive learning from both global and local perspectives to capture fine-grained inconsistencies.

    % --- LLM-based Frameworks ---
    \item \textbf{FKA-Owl}~\citep{liu2024fkaowl} is a detection model designed on MLLM, demonstrating outstanding cross-domain performance. Since FKA-Owl does not natively support fine-grained classification tasks, we fine-tuned it using the same prompts as those used for REFORM to enable comparable evaluation.

    \item \textbf{AMD}~\citep{AMD_MDSM} is a unified framework built upon MLLM designed for the MLLM-driven semantic-aligned DGM4 task. It introduces an Artifact Pre-perception Encoding module to capture manipulation traces into learnable artifact tokens and utilizes Manipulation-Oriented Reasoning to generate grounded detection results via a sequence-to-sequence format.

    % --- Agentic Framework ---
    \item \textbf{MMD-Agent}~\citep{liu2024mmfakebench} is a training-free, MLLM-based agentic framework. It decomposes the detection task into sequential sub-tasks: text-based fact-checking (retrieving external knowledge), visual manipulation analysis, and cross-modal consistency verification. To adapt this pipeline for the grounding task in ROM, we modified its visual analysis prompt to explicitly request bounding box coordinates for manipulated regions.

    % --- General Approach ---
    \item \textbf{ViLT}~\citep{ViLT_ICML} serves as the general multi-modal learning baseline. It is a representative single-stream method where cross-modal interaction layers operate on the concatenation of image and text inputs. For adaptation to the forgery detection task, we add classification and detection heads to the corresponding outputs of the model.
\end{itemize}

\subsection{Evaluation Metrics}
To comprehensively evaluate our proposed ROM, we follow the rigorous evaluation protocols and metrics outlined in~\cite{DGM4} for all manipulation detection and grounding tasks. The detailed evaluation setup is organized as follows:  

\begin{itemize}
  \item \textbf{Binary Classification.} \textbf{Accuracy (ACC)} is adopted as the evaluation metric to measure the correctness of real/fake news classification results.  

  \item\textbf{Multi-Label Classification.} 
  % For multi-label classification tasks, we employ \textbf{mean Average Precision (mAP)}. This metric effectively captures the average performance across all labels, providing a comprehensive assessment of multi-dimensional manipulation type classification accuracy.  
  For multi-label classification tasks, we employ the \textbf{mean Average Precision (mAP)}, which measures the per-class average precision and then takes the arithmetic mean across all manipulation types. This macro-averaged mAP provides a comprehensive evaluation of the model’s overall performance across different manipulation types.

  \item\textbf{Manipulated Image Bounding Box Grounding.} To evaluate the precision of predicted manipulated bounding boxes, we calculate the \textbf{mean Intersection over Union (mIoU)} between the ground-truth and predicted coordinates for all testing samples. This metric quantifies the spatial overlap between detected regions and actual manipulated areas, reflecting the localization accuracy of the model.  

  \item\textbf{Manipulated Text Token Grounding.} In the DGM4 benchmark, an additional task of manipulated text token grounding is included. For this task, \textbf{Precision ($P_{tok}$)} is used as the evaluation metric to measure the accuracy of identifying manipulated text tokens within input sequences.  
\end{itemize}

This standardized evaluation framework ensures a systematic and comparative assessment of ROM across diverse manipulation scenarios, aligning with both general detection tasks and benchmark-specific requirements.

\subsection{Task-Specific Adaptation for DGM4}
To adapt REFORM for fine-grained fake word detection on DGM4, we introduce a Token Precision Reward ($\mathcal{R}_{\text{tok}}$). 
Let $\hat{y}_{tok}$ denote the predicted manipulated words and $y_{tok}$ be the ground truth token labels. 
The reward calculates the token-level consistency:
\begin{equation}
    \mathcal{R}_{\text{tok}} = \text{ACC}(\text{Tokenize}(\hat{y}_{tok}), y_{tok}),
\end{equation}
where, $\text{Tokenize}(\cdot)$ performs text normalization (lowercasing and punctuation removal) and aligns the word-level predictions with the caption's token grid to generate a binary prediction mask. 
$\text{ACC}(\cdot)$ computes the token-wise accuracy between this mask and the ground truth. 
Crucially, for pristine samples, this metric enforces strict hallucination suppression by assigning a reward of 1.0 only if the model predicts "none" or an empty set. 
Consequently, the final total Reward when training REFORM on DGM4 is formulated as:
\begin{equation}
    \mathcal{R}_{\text{DGM4}} = \mathcal{R}_c + \mathcal{R}_a + \mathcal{R}_g + \mathcal{R}_f + \mathcal{R}_{\text{tok}}.
\end{equation}
The training settings of other stages in REFORM are consistent with the main paper.

\section{Prompt Paradigm}
\label{sec:supp_prompt}

In this section, we present the specific prompt templates designed for our proposed framework, the data distillation process, and the baseline comparisons. These templates ensure consistent task formulation across different experimental settings.

\subsection{Prompt for REFORM}
To enable fine-grained manipulation detection and localization, we design a structured prompt for the REFORM model. As shown in Fig. \ref{fig:dgm4_prompt_real}, the prompt concatenates system instructions, the news caption, and a specific set of ten options covering various manipulation types (e.g., face swap, text rewriting). Crucially, the task instruction explicitly requires the model to append the manipulated face's bounding box coordinates to the selected option if a face manipulation is detected.

\begin{figure*}[t]
    \centering
    \begin{tcolorbox}[
        colback=gray!10, 
        colframe=black, 
        boxrule=0.8pt, 
        title=\textbf{Prompt Template for REFORM}
    ]
    \small
    \textbf{[System Instruction]} \\
    \texttt{<image>} The following are multiple choice questions about fake news detection.
    
    \vspace{0.5em}
    \textbf{[Input Data]} \\
    The text caption of news is: \textit{\{News Caption\}}
    
    \vspace{0.5em}
    \textbf{[Question \& Options]} \\
    The image and text should not be manipulated. Question: Is there any manipulation in the image or text of this news? \\
    A. No. \\
    B. Image: Face swap; Text: No. \\
    C. Image: Face attribute; Text: No. \\
    D. Image: Whole generated; Text: No. \\
    E. Image: Inpainted background; Text: No. \\
    F. Image: Face swap; Text: Fully rewritten. \\
    G. Image: Face attribute; Text: Fully rewritten. \\
    H. Image: Whole generated; Text: Fully rewritten. \\
    I. Image: Inpainted background; Text: Fully rewritten. \\
    J. Image: No; Text: Fully rewritten.

    \vspace{0.5em}
    \textbf{[Task Instruction]}\\
    If the face is manipulated, locate the manipulated face in the image and append the results to your selected option. \\
    The answer is:
    The answer is:
    \end{tcolorbox}
    \caption{The prompt template for REFORM. The prompt strictly concatenates the system instruction, caption, options, and the localization instruction.}
    \label{fig:dgm4_prompt_real}
\end{figure*}

\subsection{Prompt for Reasoning Data Distillation}
To equip our model with explicit reasoning capabilities, we distill knowledge from a powerful VLM, InternVL3.5-30B~\citep{wang2025internvl3}. Fig. \ref{fig:prompt_template} illustrates the template used for reasoning generation. In this process, the ground-truth manipulation information is injected into the context, and the teacher model is instructed to generate a coherent, factual rationale explaining why the news falls into the target category, focusing on visual and textual evidence.

\begin{figure*}[t]
    \centering
    \begin{tcolorbox}[
        colback=bg_gray, 
        colframe=black, 
        boxrule=0.8pt, 
        title=\textbf{Prompt Template for Reasoning Generation}
    ]
    \small
    \textbf{[System Instruction]} \\
    You are a fake news analysis expert. You will be provided with a news item in the form of an image-text pair.
    
    \vspace{0.5em}
    \textbf{[Ground Truth Context (Dynamic)]} \\
    It is known that \textit{\{Manipulation Description\}} (e.g., "the image has been manipulated using face swap", "the image and text are original", etc.).
    
    \vspace{0.5em}
    \textbf{[Task Instruction]} \\
    Without assuming any prior knowledge about the authenticity or manipulation of this image-text news, analyze it carefully and reason only on the visual and textual evidence.
    
    \vspace{0.5em}
    \textbf{[Input Data]} \\
    The news text is: \textit{\{News Caption\}}
    
    \vspace{0.5em}
    \textbf{[Reasoning Goal]} \\
    Finally, summarize your reasoning to justify why this news can be considered that \textit{\{Target Label Description\}}.
    
    \vspace{0.5em}
    \textbf{[Constraints]} \\
    Keep your reasoning coherent, factual, and concise (\textit{\{Min Keywords\}}-\textit{\{Max Keywords\}} words). Write your response as plain continuous text, without any Markdown symbols, lists, or bullet points.
    \end{tcolorbox}
    \caption{The prompt template used to generate reasoning chains. The \textit{\{Italicized Text\}} represents dynamic content filled based on the ground-truth labels and input data.}
    \label{fig:prompt_template}
\end{figure*}

\subsection{Prompt for General-purpose Model}
For zero-shot comparisons with general-purpose MLLMs, we adapt the prompt to ensure parsing stability. As depicted in Fig. \ref{fig:prompt_strict_answer}, while the input data and options remain consistent with the REFORM setting, we modify the task instruction to request normalized coordinates (relative positioning from 0 to 1). Furthermore, explicit constraints are added to suppress intermediate reasoning and force the model to output only the final option and coordinates.

\begin{figure*}[t]
    \centering
    \begin{tcolorbox}[
        colback=bg_gray, 
        colframe=black, 
        boxrule=0.8pt, 
        title=\textbf{Prompt Template for General-purpose Model}
    ]
    \small
    \textbf{[System Instruction]} \\
    Same as REFORM \\
    \textbf{[Input Data]} \\
    Same as REFORM \\
    \textbf{[Question \& Options]} \\
    Same as REFORM \\
    \vspace{0.5em}
    
    \textbf{[Task Instruction]} \\
    If face manipulation, use rectangular box coordinates in the format of [x1,y1,x2,y2], where the top-left vertex of the image is defined as (0,0) and the bottom-right vertex as (1,1) for relative positioning, and append the results to the option you have selected.\\
    \textbf{[Constraints]} \\
    Please think carefully before giving the finnal answer, but you don't have to output the reasoning process. ONLY output the final answer in format: [Option + Coordinates (if applicable).]
    \end{tcolorbox}
    \caption{The prompt template used for General-purpose Model. Note the change in coordinate format to normalized [x1,y1,x2,y2].}
    \label{fig:prompt_strict_answer}
\end{figure*}

\subsection{Prompt for MMD-Agent}
We also evaluate the MMD-Agent baseline. Since the original MMD-Agent pipeline does not natively support grounding for specific manipulation types, we modified the prompt in Stage 2 (Image Modality) to include localization instructions. Fig. \ref{fig:mmd_agent_prompts} shows the complete workflow, where the agent sequentially performs text-based fact-checking, image manipulation detection (with our added bounding box requirement), and cross-modal consistency verification.

\begin{figure*}[t]
    \centering
    \begin{tcolorbox}[
        colback=bg_gray, 
        colframe=black, 
        boxrule=0.8pt, 
        title=\textbf{Prompt Templates for MMD-Agent Workflow}
    ]
    \small
    \textbf{[Stage 1: Fact-Checking (Text Modality)]} \\
    Given a news caption, news caption is: \textit{\{News Caption\}} \\
    Determine if there is credible objective evidence that SUPPORTS or REFUTES the news caption. Please follow the instructions below: \\
    \textbf{Thought 1:} You need to find the key entity noun in the news caption. The key entity noun could be person or object or location or event, etc. \\
    \textbf{Action 1:} Search [key entity noun]. \\
    \textbf{Observation:} \textit{\{External Knowledge / Wiki Results\}} \\
    \textbf{Thought 2:} According to Observation and other credible objective evidence, please analysis there is any objective fact that SUPPORTS or REFUTES the news caption, or if there is NOT ENOUGH INFORMATION. Analysis is: [Analysis]. \\
    \textbf{Action 2:} Draw the conclusion based on the analysis in the thought 2: if there is any credible objective evidence refuting the news caption, please answer in the form: 'Finish[TEXT REFUTES].'. If no, please answer in the form: 'Finish[TEXT SUPPORTS].'

    \vspace{0.5em}
    \hrule
    \vspace{0.5em}

    \textbf{[Stage 2: Manipulation Detection (Image Modality)]} \\
    According to the given news image, determine if the image is manipulated and identify the type of manipulation. \\
    \textbf{Classification categories:} 'Real', 'Face Swap', 'Face Expression', 'Background Swap', 'Full AI Generated'. \\
    Please follow the instructions below: \\
    \textbf{Thought 1:} Analyze the image for visual inconsistencies such as unnatural lighting, blurred edges, distorted faces, or inconsistent background noise. \\
    \textbf{Observation:} [Fact-conflicting Description] \\
    \textbf{Action 1:} Draw the conclusion based on the observation. 
    \begin{itemize}[leftmargin=1em, nosep]
        \item If Real: 'Finish[Real]'.
        \item If Face Swap/Expression: 'Finish[Face Swap, BBox: [x1, y1, x2, y2]]' (normalized 0.0-1.0).
        \item If Background/Full AI: 'Finish[Background Swap]' or 'Finish[Full AI Generated]'.
    \end{itemize}

    \vspace{0.5em}
    \hrule
    \vspace{0.5em}

    \textbf{[Stage 3: Cross-Modal Consistency (Multimodal)]} \\
    Given a multimodal misinformation, it contains both news caption and news image. News caption is: \textit{\{News Caption\}} \\
    Determine if the news caption matches the content news image. You should answer in the following forms: 'Finish[MATCH].' or 'Finish[MISMATCH].'. Please follow the instructions below: \\
    \textbf{IMAGE DESCRIPTION:} \textit{\{Image Description from Stage 2\}} \\
    \textbf{Draw the conclusion:} Based on the [IMAGE DESCRIPTION] of the news image, does the news caption match the content of news image? If yes, please answer in the form: 'Finish[MATCH].'. If no, please answer in the form: 'Finish[MISMATCH].'.
    \end{tcolorbox}
    \caption{The MMD-Agent prompt workflow. The agent sequentially performs Fact-Checking (utilizing external knowledge), Image Manipulation Detection (providing bounding boxes for faces), and Consistency Verification, before aggregating the status to determine the final authenticity.}
    \label{fig:mmd_agent_prompts}
\end{figure*}

\end{document}